\documentclass[]{fairmeta}

\usepackage{makecell}
\usepackage{amsmath}
\usepackage{booktabs}
\usepackage[table]{xcolor}
\usepackage{amsmath}
\usepackage{amssymb}
\usepackage{subcaption}
\usepackage{tcolorbox}
\usepackage{wrapfig}
\usepackage[commandnameprefix=always]{changes}
\usepackage{multirow}
\usepackage{booktabs}

\usepackage{times}
\usepackage{latexsym}

\usepackage[T1]{fontenc}

\usepackage[utf8]{inputenc}

\usepackage{microtype}

\usepackage{inconsolata}

\usepackage{graphicx}

\title{Scaling Reinforcement Learning for Content Moderation with Large Language Models}

\author[]{Rui Liu}
\author[]{Yuchen Lu}
\author[]{Zhenyu Hou}
\author[]{Fangzhou Xiong}
\author[]{Xiaoyang Zhang}
\author[]{Changshu Jian}
\author[]{Zhicheng Zhu}
\author[]{Jiayuan Ma}
\author[]{Jacob Tao}
\author[]{Chaitali Gupta}
\author[]{Xiaochang Peng}
\author[]{Shike Mei}
\author[]{Hang Cui}
\author[]{Yang Qin}
\author[]{Shuo Tang}
\author[]{Jason Gaedtke}
\author[]{Arpit Mittal}
\author[]{Hamed Firooz}

\affiliation[]{Meta AI}

\newcommand{\phishing}{{Task1~}}
\newcommand{\ubp}{{Task2~}}
\newcommand{\hpi}{{Task3~}}

\if 1
    \newcommand{\hamed}[1]{{\color{red}Hamed: #1}}
    
\else
    \newcommand{\hamed}[1]{}
    \newcommand{\yuchen}[2]{}
\fi

\abstract{
Content moderation at scale remains one of the most pressing challenges in today's digital ecosystem, where billions of user- and AI-generated artifacts must be continuously evaluated for policy violations. While recent advances in large language models (LLMs) have demonstrated remarkable potential for policy-grounded content moderation, the practical challenges of training these systems to achieve "expert-level" accuracy in real-world moderation scenarios remain largely unexplored, a domain characterized by label sparsity, evolving policy definitions, and the critical need for nuanced reasoning beyond shallow pattern matching. In this work, we present a comprehensive empirical investigation into scaling reinforcement learning (RL) for content classification and systematically evaluate multiple RL training recipes and reward shaping strategies, including \emph{verifiable rewards} and \emph{LLM-as-judge} frameworks, to transform general-purpose language models into specialized, policy-aligned classifiers across three real-world content moderation tasks.
Our findings reveal actionable insights for industrial-scale content moderation systems. Most notably, we demonstrate that RL exhibits \emph{sigmoid-like scaling behavior} where performance improves smoothly with increasing training data, number of rollouts, and optimization steps, before gradually saturating. In addition, we show that RL substantially improves performance on tasks requiring complex, policy-grounded reasoning, while achieving up to \(100\times\) higher data efficiency than supervised fine-tuning (SFT). This makes RL particularly effective in domains where expert annotations are scarce or costly.
}

\date{\today}
\correspondence{Hamed Firooz at \email{mhfirooz@meta.com}}

\begin{document}
\maketitle

\section{Introduction}
In today’s digital ecosystem, content moderation has become a foundational requirement for maintaining safe, trustworthy, and policy compliant online environments. Search and AI platforms such as Google \citep{google_text_moderation_2023,weidinger2024holistic}, Social Networks such as Meta’s Facebook and Instagram \citep{meta_connect2024_responsible_approach,inan2023llamaguard}, online retail such as Amazon \citep{gampa2023prioritised}, user-facing AI chat such as OpenAI \citep{openai_moderation,guan2024deliberative, markov2023holistic}, and Anthropic \citep{anthropic_constitutional_ai} all host or generate vast volumes of user- and AI-created text, images, and videos at global scale. As the boundaries between user-generated and model-generated content blur, these organizations face parallel challenges in detecting, scoring, and mitigating harmful or policy-violating material. Without effective moderation, these ecosystems risk the rapid spread of misinformation, harassment, hate speech, and other harms that undermine user safety, advertiser trust, and regulatory compliance \citep{gao2025cannot}. Consequently, modern moderation pipelines increasingly combine machine learning classifiers, human review, and adaptive policy frameworks to ensure that content decisions remain both scalable and aligned with evolving community and societal standards.

Large Language Models (LLMs) have recently emerged as a promising paradigm for improving content moderation due to their strong linguistic reasoning and generalization capabilities \citep{markov2024systems, yuan2025hard}. By leveraging rich semantic representations learned during large-scale pretraining, LLMs can capture subtle contextual cues, linguistic nuances, and domain-specific patterns that traditional content classifiers often miss. Previous work shows that LLM can act as rule-following moderators via prompting \citep{kumar2024watch}, as supervised fine-tuned (SFT) classifiers \citep{ma2024adaptinglargelanguagemodels} or as guardrail models such as Llama Guard \citep{inan2023llamaguard} and BingoGuard \citep{yin2025bingoguard}, with industry adoptions such as Google Ads’ LLM-based content review system \citep{qiao2024scaling}. 

These works mainly focus on how to adapt LLMs to moderation tasks through prompting or supervised fine-tuning, enabling models to follow predefined rules and classify content into safety taxonomies. 
However, while effective for many scenarios, prompting and SFT still face the challenge to encode highly complex, conditional, and context-dependent moderation policies. 
Real-world content moderation policies often involve hierarchical severity levels, exception clauses, and nuanced distinctions that depend on subtle linguistic cues or multi-turn context \citep{sharma2022detecting}. Capturing these behaviors through static supervision alone can lead to inconsistencies, overfitting to annotation artifacts, or poor generalization in ambiguous or adversarial cases.

In parallel, reinforcement learning (RL) has emerged as a post-training technique for aligning general-purpose LLMs with human preferences, particularly in safety-critical settings. 
Prior work—including Constitutional AI \citep{sharma2025constitutional}, Safe-RLHF \citep{dai2023safe}, deliberative alignment \citep{guan2024deliberative}, RealSafe-R1 \citep{zhang2025realsafe}, and RigorLLM
\citep{yuan2024rigorllm}—demonstrates that RL can optimize behaviors that extend beyond token-level supervised training. 
In these systems, RL allows models to reason over and internalize complex safety policies, integrate multi-step constraints, and balance competing behavioral requirements that are difficult to encode through SFT alone \citep{mu2024rule, mu2024ruleicml}. However, despite RL’s demonstrated success in LLM safety alignment, it has not yet been scaled or systematically applied to content moderation in large scale industrial setting, a domain where nuanced policy interpretation, hierarchical taxonomies, and fine-grained reasoning are especially crucial.

Motivated by this gap, we focus in this work on scaling RL for post-training LLMs for content moderation in products offered by \textsc{Meta Platforms, Inc}. We investigate multiple training recipes and reward shaping strategies, including \emph{verifiable rewards} (RLVR) and \emph{LLM-as-judge} setups. 
We show how RLVR is challenging to apply directly to content moderation task because many safety and policy definitions do not admit verifiable ground-truth rewards and they are susceptible to reward hacking. 
As a result, our recipe rely on reward shaping combining verifiable reward, rubric-based evaluators, and LLM-judge rewards, which provide structured, policy-aligned feedback in addition to binary verifiable signals, enabling RL to operate effectively even when correctness cannot be determined through automated verification.

We evaluate our methods across three \textsc{Meta Platforms, Inc.} policy-violation classification tasks derived from real-world production data. We further discuss the challenges of scaling reinforcement learning systems for industrial content moderation and present a practical training recipe that guides key optimization and data-allocation choices when adapting general-purpose large language models (LLMs) into specialized, policy-aligned classifiers capable of surpassing average human performance and approaching expert-level accuracy. In summary, our key findings are as follows:

\begin{enumerate}
    \item \textbf{Effective mitigation strategies for critical RL failure modes.}  
    Across tasks, we observe several characteristic challenges in RL post-training, including bi-polar (bimodal) probability distributions, reward hacking, and length-collapse effects that obscure both faithfulness and factuality in RL-trained content moderation models. We analyze how these phenomena emerge during optimization and introduce concrete interventions—such as rubric-based rewards and Monte-Carlo–based score aggregation—that substantially stabilize training and improve model robustness.
    
    \item \textbf{$10\times$--$100\times$ higher data efficiency compared to SFT.}  
    Across tasks, RL-Only models trained on a few hundred examples often match or surpass the performance of SFT models trained on tens of thousands of labeled samples. This makes RL particularly attractive in domains such as content moderation, where high-quality labels are expensive and time-consuming to obtain. Moreover, we observe that large-scale SFT can overly constrain the model’s behavior and hinder exploration during the subsequent RL stage, whereas RL-Only training on a base model avoids this issue and maintains broader exploration capacity.

    \item \textbf{Predictable scaling behavior with both data and compute.}  
    We show that RL follows sigmoid-like scaling trends: performance improves smoothly with additional training data, number of rollouts, and optimization steps, and then gradually saturates. This provides a practical blueprint for allocating compute and designing rollout budgets in real-world RL pipelines.

    \item \textbf{Rubric-Based Reasoning Reward tailored for content moderation.}  
    Our reward model evaluates the entire reasoning trace using policy-grounded qualitative criteria, enabling fine-grained supervision beyond the final label. We further analyze the role of reward shaping—combining accuracy, rubric, format, and length rewards—and provide empirical evidence that shaped rewards substantially improve model faithfulness, consistency, and downstream performance.
\end{enumerate}

\section{Setup}
\subsection{Prompt}\label{ss:prompt}


Content moderation can be posed as a classification problem: given a piece of content $\mathcal{C}$ and a policy $\mathcal{P}$, we aim to estimate the probability that $\mathcal{C}$ violates $\mathcal{P}$. Let the input prompt, $q$, be
\begin{equation}
q = \mathrm{concat}\!\big(\texttt{Instruction: } \mathcal{I};\\
                    \texttt{ Policy: } \mathcal{P};\\ 
                    \texttt{ Content: } \mathcal{C}\big)
\end{equation}
where $\mathcal{I}$ is the task instruction. The model induces a conditional distribution over labels $y \in \{y_1,\dots,y_K\}$ defined in the policy,
\begin{equation*}
P_\theta(y = y_i\mid q), \quad i=1,\dots,K.    
\end{equation*}

In this work we focus on binary classification ($K=2$; i.e. violation vs non-violation), where we take $y=1$ to indicate a policy violation and $y=0$ otherwise.

\begin{tcolorbox}[
    title=Example Input Prompt, 
    colback=gray!10, 
    colframe=black, 
    fontupper=\small\raggedright,
    fonttitle=\small
]
Imagine you are reviewing for [CONTENT\_MODERATION\_TASK]. Given the following content, determine the answer.\\[1em]

Content: [CONTENT\_MULTIMODAL]
Account information:\\[1em]
User/Account name: [USERNAME],\\
Bio: [BIO\_TEXT].\\
Account stated country: [COUNTRY\_CODE],\\
Follower count: [NUMBER],\\
Recent contents: [CONTENT\_MULTIMODAL\_1] \\[1em]

The review process should follow the following step:

Step 1. [POLICY\_1] [POLICY\_1\_CONTENT].

Step 2. [POLICY\_2] [POLICY\_2\_CONTENT].

Step 3: [CONCLUSION] Provide final assessment.\\[1em]

Please think step-by-step then provide your independent assessment in JSON format, including the following keys:

[KEY\_1]: for [POLICY\_1] (true/false)\newline
[KEY\_2]: for [POLICY\_2] (no/yes)\newline
[KEY\_3]: for [CONCLUSION] (true/false)

\end{tcolorbox}\label{input_prompt}

\begin{tcolorbox}[
    title=Example Model Generation, 
    colback=gray!10, 
    colframe=black, 
    fontupper=\small\raggedright,
    fonttitle=\small
]
<think>Okay, let's start by going through each step carefully.\\
  **Step 1: ...\\
  **Step 2: ...\\
  **Step 3: ... \\
  **Final Decision: Since ... is false.\\
  </think>\\[1em]
  
\{\\
\quad [KEY\_1]: "false",\\
\quad [KEY\_2]: "yes",\\
\quad [KEY\_3]: "true"\\
\}
\end{tcolorbox}

The model outputs a structured triple
\begin{equation}\label{eq:output}
(r, \hat{y},\ \hat{p}),
\end{equation}
where $\hat{y} \in \{1, 0\}$ is the predicted label in language space (violation / non-violation), $\hat{p} = P_\theta(\hat{y}=1 \mid q)$ is the associated probability to that label in language space, and $r$ is the chain-of-thought reasoning of the model rationale to get to label $\hat{y}$~\citep{wei2022chain}. At inference time we are interested in probability of class $y_i$ given the context $P_\theta (y=y_i \mid q)$ and apply threshold $\tau$
\begin{equation}
\hat{y} = \mathbb{I}\!\big[P_\theta(y=1 \mid q) \ge \tau\big],   
\label{eq:thresholding}
\end{equation}
with threshold $\tau \in [0,1]$ calibrated on a validation set to satisfy precision/recall or cost-sensitive targets.




\subsection{Experimentation Setup}
For our RL algorithm, we use Group Relative Policy Optimization (GRPO)~\citep{shao2024deepseekmath}, a more compute-efficient alternative to Proximal Policy Optimization (PPO)~\citep{schulman2017proximal}. GRPO eliminates the need for an explicit value function by computing relative advantages across a group of sampled responses. Given a prompt $q$, we draw $N$ rollouts $\{o_1, \ldots, o_N\}$ from the current policy $\pi_\theta$, obtain their scalar rewards $\{R_1, \ldots, R_N\}$, and compute group-normalized advantages:
\[
A_i = \frac{R_i - \mu_R}{\sigma_R + \epsilon},
\]
where $\mu_R$ and $\sigma_R$ denote the mean and standard deviation of the group rewards, respectively. The GRPO objective is then defined as:
\[
\mathcal{L}_{\text{GRPO}}(\theta) 
=
\frac{1}{N}
\sum_{i=1}^{N}
\operatorname{clip}\!\left(
\frac{\pi_\theta(o_i \mid q)}{\pi_{\theta_{\text{old}}}(o_i \mid q)},
1 - \epsilon,\, 1 + \epsilon
\right)
A_i
\;-\;
\beta \,
\mathrm{KL}\!\left(
\pi_\theta(\cdot \mid q)
\;\|\;
\pi_{\theta_{\text{ref}}}(\cdot \mid q)
\right),
\]
which updates the policy toward higher-quality samples identified through groupwise comparison. This relative-feedback formulation avoids value estimation and improves optimization stability in settings with sparse or noisy reward signals.

In all experiments, we set the KL coefficient $\beta = 0$, following recommendations from \citep{liu2025understanding, shao2024deepseekmath}. Empirically, removing the KL penalty yields consistently better performance in our content moderation tasks, as the policy benefits from stronger exploration without collapsing toward the initial SFT distribution.

We further apply sequence-level normalization of rewards, rather than token-level normalization, which we find substantially improves training stability and final model quality. This design choice aligns with guidance from GSPO~\citep{zheng2025group}, where sequence-level normalization preserves the relative ordering of complete trajectories and yields more reliable optimization dynamics for reasoning-intensive tasks.

\subsection{Frameworks}
We evaluated our reinforcement learning pipeline using both HuggingFace TRL~\citep{vonwerra2020trl} and Verl~\citep{verl2024hybridflow} in order to determine which framework provides higher end-to-end training throughput. We define throughput as total number of tokens (input + output) processed by a single GPU per second:
\begin{equation}
\text{Throughput} = \frac{\text{Total tokens processed}}{\text{Number of GPUs} \times \text{Time}},
\end{equation}
where $\text{Time}$ is measured in seconds.

As shown in Table~\ref{tab:throughput_comparison}, Verl, due to its HybridFlow-based execution backbone, consistently achieves substantially higher throughput than TRL, reaching up to $2.5\times$ improvement under comparable settings on both internal tasks and external benchmark dataset GSM8K \citep{cobbe2021training}.

\begin{table*}[ht]
\centering
\small
\begin{tabular}{llrrl}
Dataset & Model & \multicolumn{2}{c}{Training Efficiency (tokens/s/GPU)} & VeRL vs TRL \\
\cline{3-4}
 &  & VeRL & TRL &  \\
\Xhline{3\arrayrulewidth}
\ubp   & Qwen2p5 VL 7b & 4600 & 1854 & 2.5x \\
\hline
GSM8K & Qwen2p5 VL 7b & 1500 &  730 & 2.0x \\
\hline
\end{tabular}
\caption{Training efficiency comparison of VeRL vs TRL (tokens/s/GPU) for Qwen2p5 VL 7b on \ubp and GSM8K.}
\label{tab:throughput_comparison}
\end{table*}

\section{Challenges}
In this section, we analyze the key challenges that arise when applying RL to real-world content moderation tasks, with a particular focus on factors that limit the scalability, stability, and quality of RL-based training. Our discussion centers on challenges encountered under a reinforcement learning setup driven by \emph{verifiable rewards}, where supervision is derived primarily from objective final-label correctness. While such reward is attractive due to its simplicity, it introduce unique difficulties related to reward design, verification, and optimization dynamics. Through a series of empirical analyses, we highlight how these challenges manifest in practice and motivate the design choices introduced in subsequent sections.

\subsection{Data and Label Scarcity}
Data scarcity remains a major operational barrier in content moderation due to the time and cost required to acquire large-scale, high-quality expert labels \citep{kiela2020hateful, alam2022survey}. In practice, annotation workflows begin with policy makers labeling a few hundred representative examples to define the ground-truth standard. Policy teams then train expert reviewers, a ramp-up process that typically takes couple of months. Once onboarded, expert reviewers generate on the order of a few hundreds labels per week.

Despite this investment, expert labels often require multiple rounds of review, feedback, and correction, with each iteration adding roughly several weeks of latency. Consequently, scaling from a few hundred seed labels to a few thousand high-quality expert labels usually takes several months and results in at least a tenfold increase in human-labeling costs.

\subsection{Verification and Reward-Design}\label{sec:VerificationAndRewardDesign}
\subsubsection{Lack of a Verification Process}
Content moderation requires forms of reasoning similar to those used in tasks such as code generation or mathematical problem solving, where models must navigate complex, rule-driven decision spaces. A key difference, however, is that content moderation lacks a reliable mechanism for verifying intermediate reasoning steps. In coding, compilers provide deterministic feedback on syntactic and logical correctness, and in mathematics, intermediate steps can often be symbolically checked~\citep{jha2024rlsf}. By contrast, content moderation has no analogous “safety compiler’’ that can systematically audit or validate a model’s chain-of-thought.

This absence of intermediate verification poses a major challenge for training and evaluation. In Section~\ref{ss:reflection-aided}, we show that reflection mechanisms allow the model to self-evaluate its reasoning trace and revise its conclusions. Complementarily, Section~\ref{s:reward_shaping} demonstrates how rubric-based rewards can be used to assess and improve the overall quality of reasoning, providing a practical substitute for explicit step-by-step verification in moderation tasks.

\subsubsection{Susceptible to Reward Hacking}
\begin{wrapfigure}{r}{0.48\linewidth}
    \vspace{-10pt}
    \centering
    \includegraphics[width=1\linewidth]{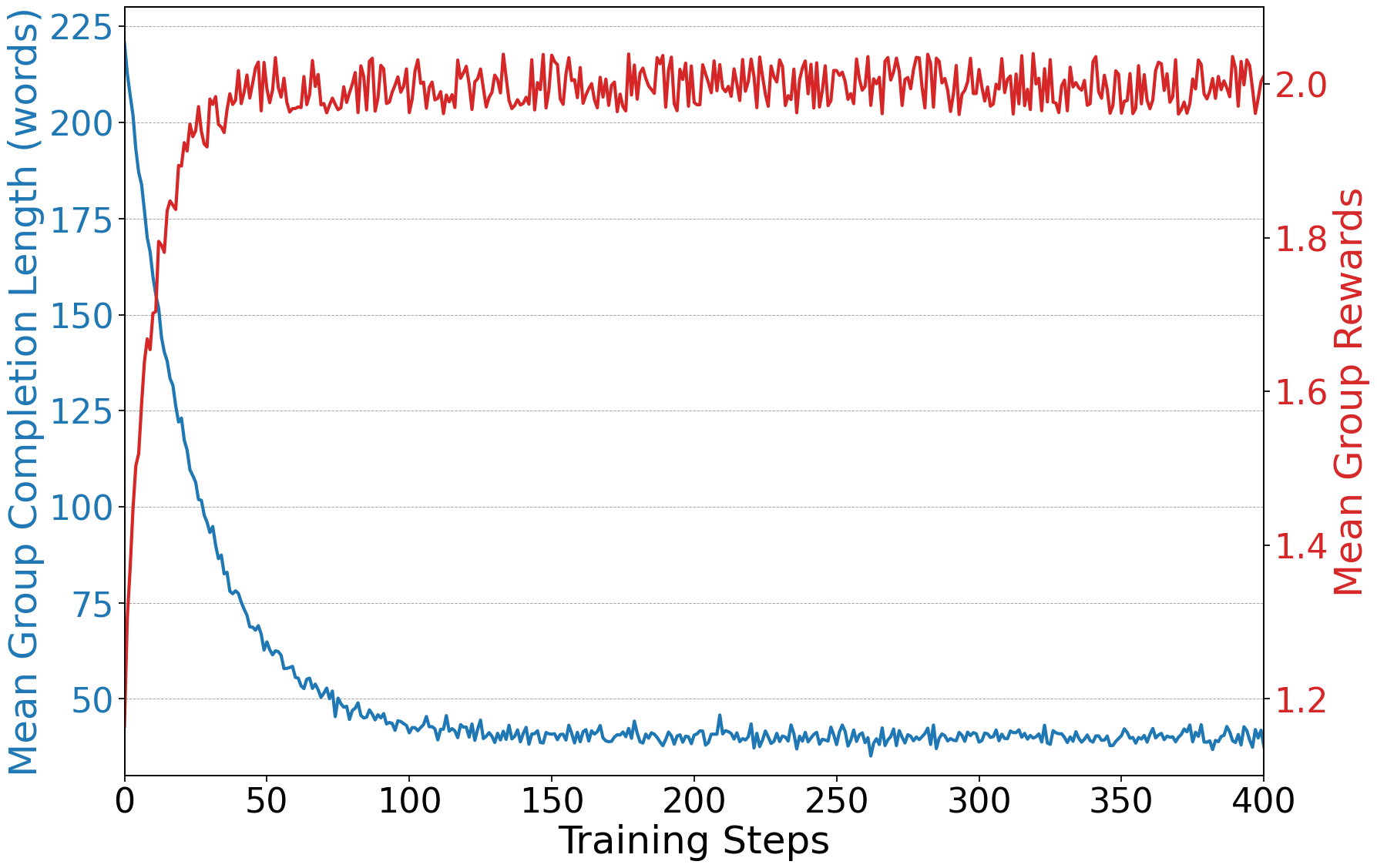}
    \caption{Accuracy-based rewards induce reward hacking: explanation length collapses over training, and responses degenerate into short label assertions.}
    \label{fig:accuracy_reward_hack}
    \vspace{-15pt}
\end{wrapfigure}
For our content-moderation tasks, we find that simple, verifiable rewards based on final-label matching ($R_{\text{acc}}$ in Eq.~\eqref{eq:rewards}) quickly reach a performance ceiling. 

Figure~\ref{fig:accuracy_reward_hack} shows a typical trajectory in which the model’s reasoning length steadily collapses (from roughly 250 words to fewer than 50), yielding extremely brief or semantically empty explanations followed by a bare True/False prediction. This behavior indicates that the model is not learning deeper task structure or producing meaningful reasoning chains; instead, it is exploiting the reward by shortcutting directly to the final label. To reduce length-dependent bias in the GRPO updates, we apply sequence-level reward normalization, following the stabilization strategy proposed in GSPO~\citep{zheng2025group}, which improves optimization behavior for long-form reasoning tasks.

\subsubsection{Factuality and Faithfulness}
We observe a distinct trade-off between two types of hallucination during RL optimization: 
(1) \textbf{faithfulness}, which measures the model's ability to follow instructions, and 
(2) \textbf{factuality}, which measures the model's adherence to the factual information specified in the policy. 
To evaluate both faithfulness and factuality of the trained policy model, we use the state-of-the-art LLM-based judge Gemini-2.5-Pro~\citep{comanici2025gemini}. 
For factuality in particular, we further leverage the Hughes Hallucination Evaluation Model (HHEM)~\citep{hughes2023hhem}.

As training progresses, RL-optimized policies often appear to improve instruction following~\citep{huang2024survey}, while their measured factuality error rate decreases. However, this apparent improvement is largely an artifact of \emph{length collapse}: the policy learns to produce increasingly short outputs—an effect illustrated in Figure~\ref{fig:accuracy_reward_hack}—which reduces the number of explicit factual statements and therefore the number of opportunities for detectable errors. In practice, the model’s underlying grounding does not improve. Instead, the policy increasingly generates post-hoc rationales crafted to align with the ground-truth label, rather than engaging in genuine, input-grounded reasoning.

\begin{figure}[ht]
    \centering
    \begin{subfigure}[t]{0.48\textwidth}
        \centering
        \includegraphics[width=\textwidth]{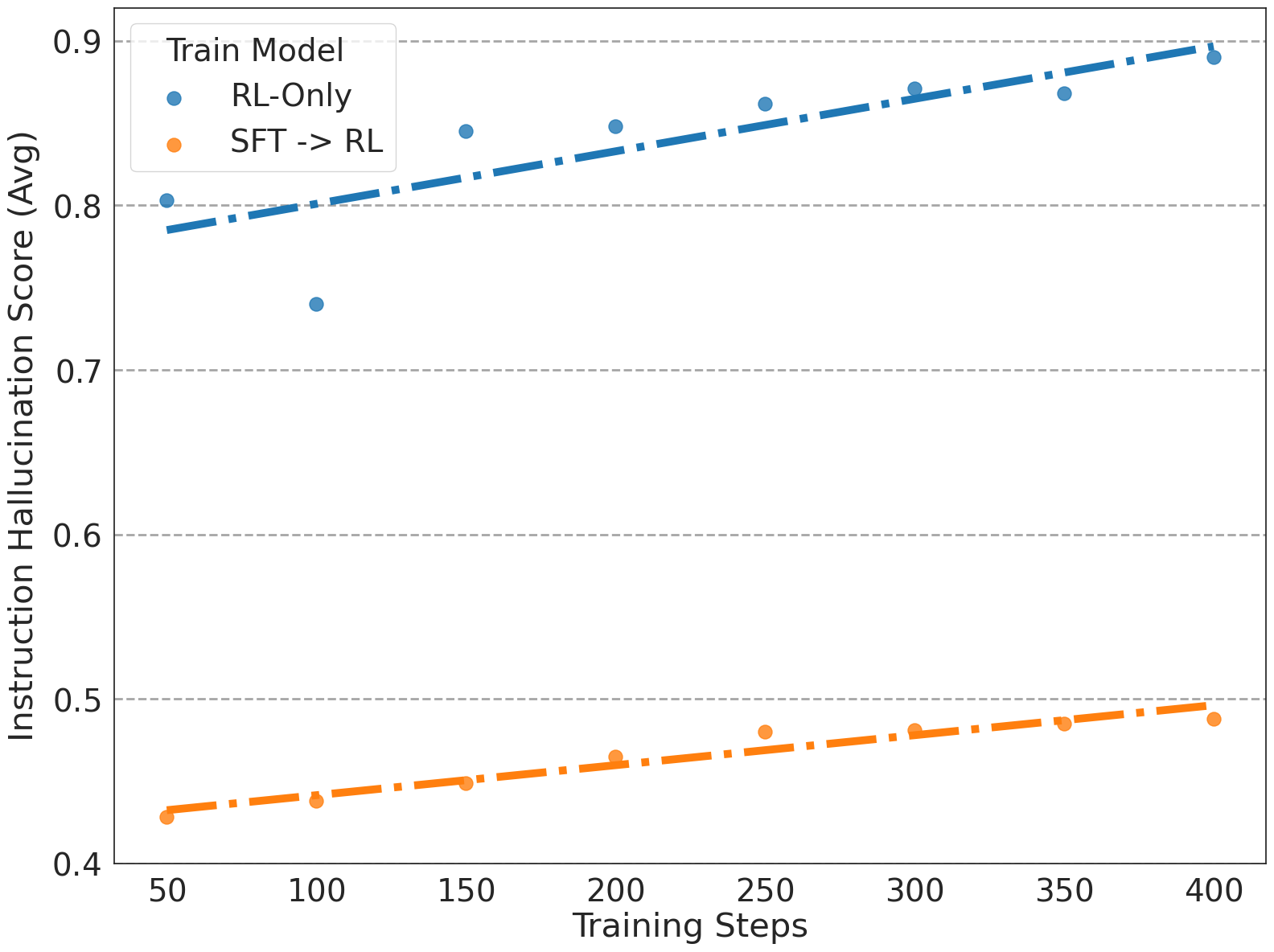}
        \caption{Faithfulness: Instruction hallucination score at each step}
        \label{fig:instruction_halu_a}
    \end{subfigure}
    \hfill
    \begin{subfigure}[t]{0.48\textwidth}
        \centering
        \includegraphics[width=\textwidth]{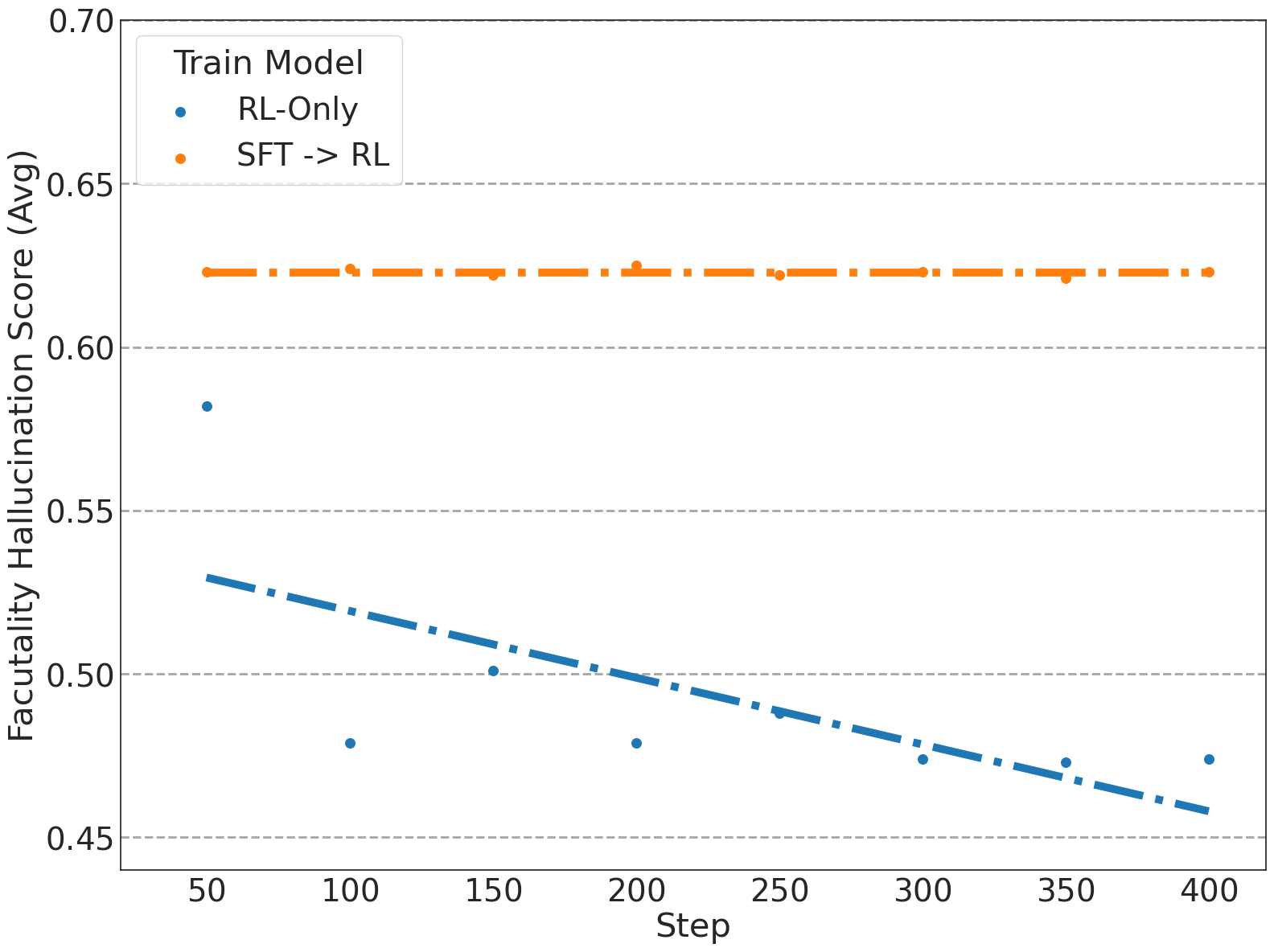}
        \caption{Factuality hallucination score at each step}
        \label{fig:factuality_halu_b}
    \end{subfigure}
    \caption{Faithfulness and factuality under RL ablations. “Instruction hallucination” and “factuality hallucination” are the complements of faithfulness and factuality, respectively. }
\end{figure}

To further examine these optimization dynamics, we compare two common training recipe for LLM-based classifiers: 
\paragraph{(a) Direct RL (RL-Only).}
Applying RL directly to the base model leads to a severe degradation in \emph{instruction adherence}, manifesting as increased instruction hallucination, as shown in Figure~\ref{fig:instruction_halu_a}. Without the inductive bias provided by supervised demonstrations, the policy fails to converge to a stable reasoning schema and instead produces unstructured, off-topic, or constraint-violating outputs. This behavior reflects classical under-regularization in policy optimization, in which the model exploits reward shortcuts rather than learning task-consistent reasoning patterns.

\paragraph{(2) Two-Stage Training (SFT $\rightarrow$ RL).}
Initializing with SFT substantially stabilizes RL, anchoring the model in a coherent reasoning structure that prevents the degenerate behaviors observed in the RL-Only setting. However, this regularity introduces a secondary issue: a higher incidence of \emph{factuality hallucinations}, as shown in Figure~\ref{fig:factuality_halu_b}. Under optimization pressure, the model generates plausible, well-structured rationales crafted to support the ground-truth label, even when these explanations lack semantic correctness or grounding in the content moderation policy specified in the prompt. Thus, although the reasoning format remains intact, the underlying factuality does not improve—and may in fact degrade.

In our experiments, we measure "instruction hallucination"—that is the complement of faithfulness: higher faithfulness indicates fewer invented or misinterpreted instructions. For factuality, we measure "factuality hallucination" which is reported as the complement of factuality: higher factuality indicates that the model’s reasoning remains grounded in evidence and avoids introducing unsupported or incorrect claims.

\subsection{Bi-polar probability distribution}

Following Eq.~\eqref{eq:output}, for reasoning models, a generation given the input prompt $q$ is $o=(r, \hat{y}, \hat{p})$. In this case, the token probability, $P_\theta(y\mid r, q)$, is a conditional probability dependent on both $r$ and input prompt $q$, and we observe it is more bimodal in classification use cases (as shown in Figure~\ref{fig:monte_carlo_dist_diff}) because the conclusion about the correct class is often already available in the reasoning trace (see prompt example in Section~\ref{ss:prompt}). This bimodal score distribution leads to poor score-based performance (PRAUC and R@P90) that requires calibration techniques or alternative confidence estimation methods to improve score discrimination between correct and incorrect predictions. 

To address this bimodality, we investigate two complementary mitigation strategies. First, in Section~\ref{ss:monte-carlo}, we smooth the score distribution by aggregating probabilities over multiple sampled reasoning traces via a Monte-Carlo estimator, which provides a more calibrated approximation of $P_\theta(y\mid q)$. Second, in Section~\ref{ss:reflection-aided}, we introduce a reflection-aided prompting scheme that encourages the model to revisit its initial decision before producing a final label, yielding better-behaved confidence scores and reducing extreme polarization in the output distribution.

\subsubsection{Monte-Carlo method}\label{ss:monte-carlo}
\begin{wraptable}{r}{0.42\linewidth}
    \vspace{-10pt}
    \centering
    \begin{tabular}{cc}
        \hline
        N & Gain on R@P90 \\
        \hline
        1 & 0.03 \\
        3 & 0.04 \\
        4 & 0.05 \\
        5 & 0.05 \\
        8 & 0.03\\
        16 & 0.05\\
        32 & 0.05 \\
        \hline
    \end{tabular}
    \caption{Performance gain of \ubp after applying Monte-Carlo sampling at T = 1.0 for different N, compared to baseline of T = 0.0 and N = 1}
    \vspace{-25pt}
    \label{tab:monte_carlo_gain_on_n}
\end{wraptable}

Our solution is to estimate the probability score through the Monte-Carlo method, which samples a sufficient number of responses, and approximates the overall probability. With the law of total probability, we have the following equation:
\begin{align*}
P_\theta(y\mid q) &= \sum_{r} P_\theta(y\mid r , q) P(r\mid q) \\
&= \mathbb{E}_{r \sim P_\theta(r\mid q)}[P_\theta(y\mid r, q)]
\end{align*}

The probability of output $y$ given $q$ is the expected conditional probability of $y$ given $q$ and a sampled reasoning trace $r$ (COT), weighted by the likelihood of each reason. This approach helps in overcoming the challenges posed by the bi-polar probability distribution observed in reasoning models by providing a more robust estimation through comprehensive sampling of the thought space.

\begin{figure}[ht]
    \centering
    \begin{subfigure}[t]{0.48\textwidth}
        \centering
        \includegraphics[width=\textwidth]{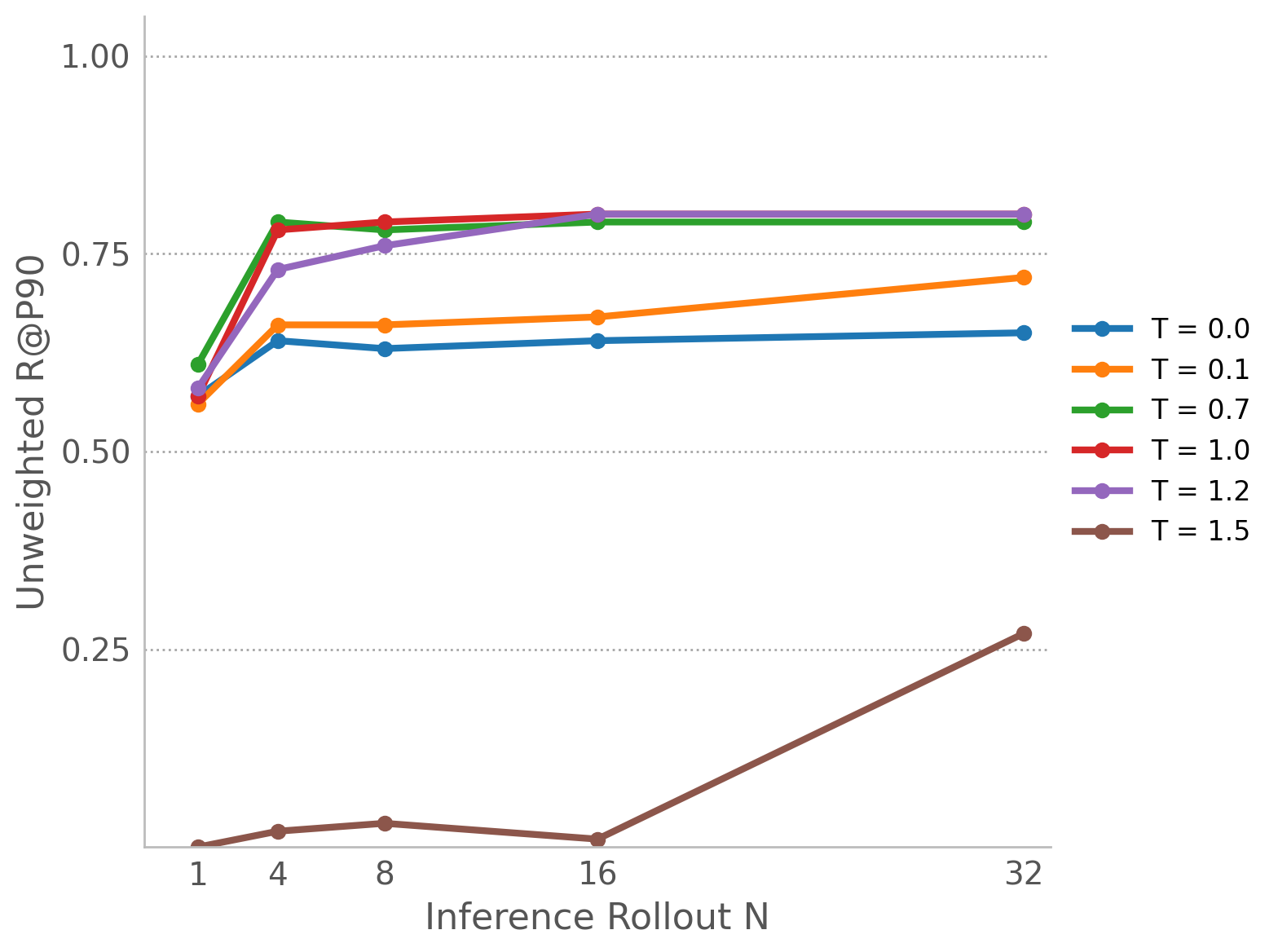}
        \label{fig:phishing_mc}
    \end{subfigure}
    \hfill
    \begin{subfigure}[t]{0.48\textwidth}
        \centering
        \includegraphics[width=\textwidth]{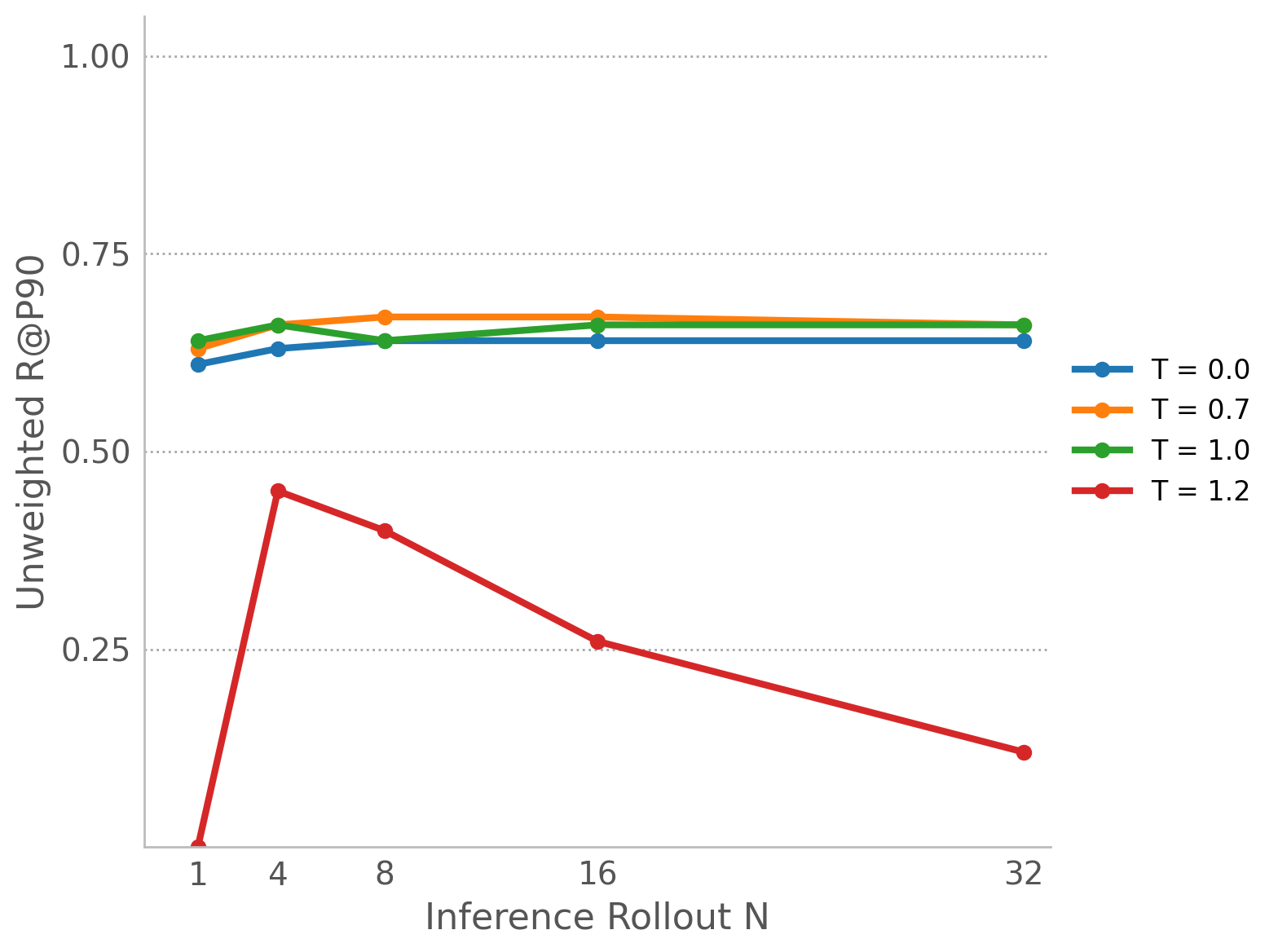}
        \label{fig:ubp_mc}
    \end{subfigure}
    \caption{Monte-Carlo Sampling at Different N (number of rollouts) and T (temperature) for \phishing and \ubp}
    \label{fig:monte_carlo_different_n_t}
\end{figure}

We tune two primary hyperparameters in the Monte Carlo sampling procedure: the number of rollouts $N$ and the sampling temperature $T$. Figure~\ref{fig:monte_carlo_different_n_t} illustrates how varying $N$ and $T$ affects model performance.

\begin{enumerate}
    \item \textbf{Effect of the number of rollouts.}  
    At moderate sampling temperatures ($T \leq 1.0$), increasing the number of rollouts $N$ consistently improves performance, albeit with diminishing returns. Table~\ref{tab:monte_carlo_gain_on_n} reports the performance gains obtained by Monte Carlo sampling at $T = 1.0$ for different values of $N$. We observe that performance plateaus beyond $N = 4$, indicating that test-time scaling~\citep{muennighoff2025s1} can be effective.
    
    \item \textbf{Effect of sampling temperature.}  
    We find that the optimal sampling temperature lies between $0.7$ and $1.0$. For temperatures above $1.0$, performance degrades due to an increased incidence of parsing errors and generation anomalies, which negatively impact downstream evaluation.
\end{enumerate}

\begin{figure}
    \centering
    \includegraphics[width=0.7\linewidth]{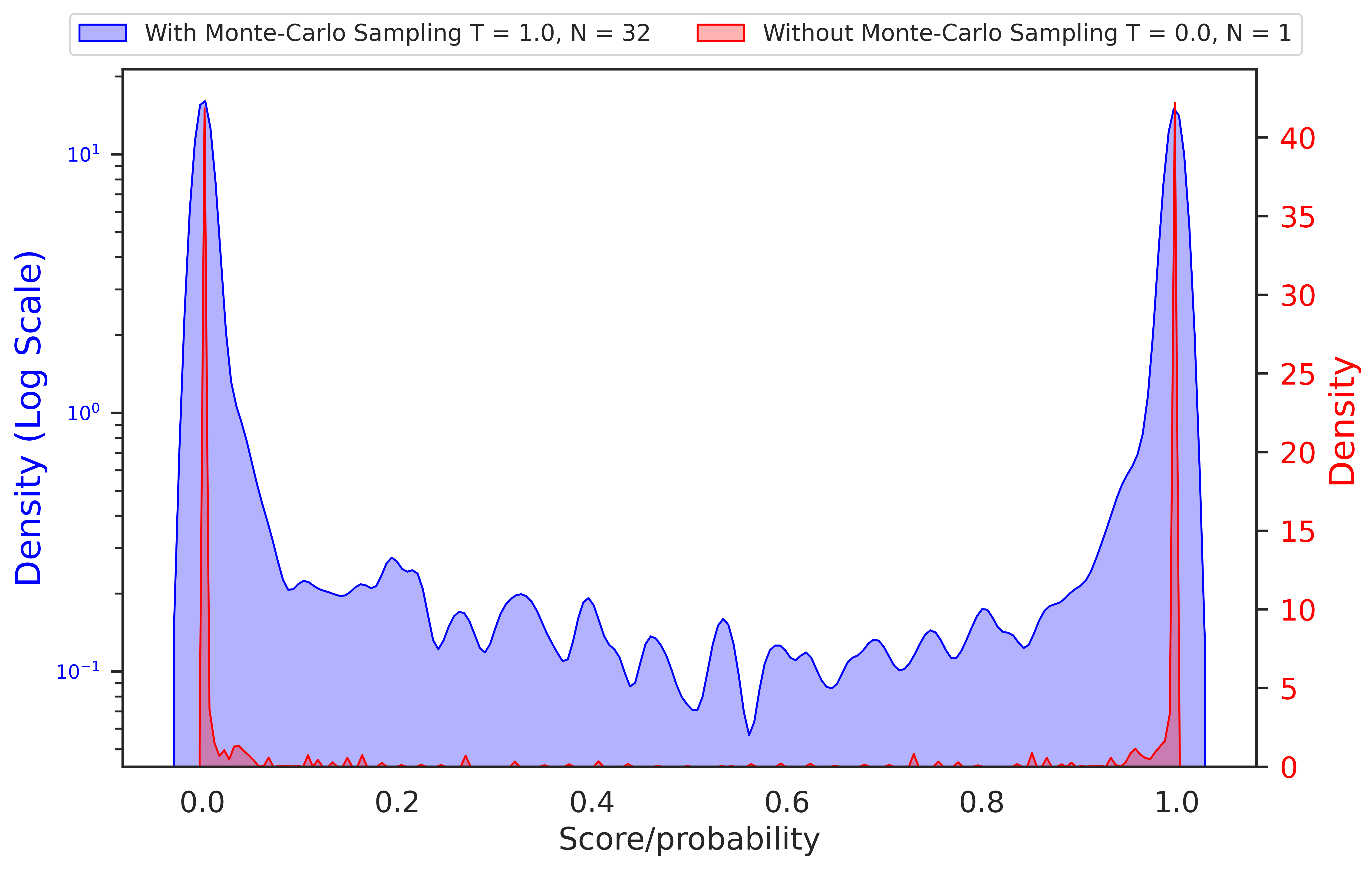}
    \caption{Comparison of label token probability distribution: without Monte-Carlo sampling vs with Monte-Carlo sampling}
    \label{fig:monte_carlo_dist_diff}
\end{figure}

To confirm how Monte-Carlo sampling helps with mitigating bi-polar probability distribution, we compare the label token distribution with and without our sampling strategy in Figure \ref{fig:monte_carlo_dist_diff}. We can observe that such strategy indeed shifts some of the extreme probability mass into the center to account for model uncertainty. 

\subsubsection{Reflection-aided Prompting}\label{ss:reflection-aided}

Inspired by work in literature about LLM reflection during reasoning \citep{shinn2023reflexion}, we leveage a three-stage prompting strategy for binary classification after model thinking: the model (i) emits an initial label (the \emph{first decision}), (ii) reflects on evidence via sub-labels (e.g. "was there any URL exist in the content"), and (iii) outputs a final label. This design is motivated by the observation that the log-probability of the final label token is often extremely polarized, which exacerbates the thresholding difficulty in Eq.~\eqref{eq:thresholding}. By asking the model to reflect before issuing a final label, we obtain better-behaved score distributions. Our prompting template is shown in Table~\ref{tab:conclusion_reflection}.

\begin{table}[ht]
\begin{tcolorbox}[
    title=Example Model Generation for Classification with reflection template, 
    colback=gray!10, 
    colframe=black, 
    fontupper=\small\raggedright,
    fonttitle=\small
]
<think>Okay, let's start by going through each step carefully.\\
  **Step 1: ... sub\_label\_1 is false\\
  **Step 2: ... sub\_label\_1 is true\\
  **Step 3: Final Decision**
  Since ... is false.\\
  </think>\\[1em]
  
\{\\
first\_decision": true/false,\\
"sub\_label\_1": true/false,\\
"sub\_label\_2": true/false,\\
...\\
"last\_decision": true/false\\
\}
\end{tcolorbox}
\caption{Prompt template and example model output for reflection-aided classification.}
\label{tab:conclusion_reflection}
\end{table}

\begin{wraptable}{r}{0.42\linewidth}
\vspace{-10pt}
\centering
\caption{Performance of LLM moderator with and without reflection for \hpi}
\label{tab:first_last_decision_scoring}
\begin{tabular}{ccc}
\hline
Scoring method & PRAUC & R@P90 \\
\hline
Without Reflection   & 0.77 & 0.05 \\
With Reflection  & \bf{0.89} & \bf{0.59} \\
\hline
\end{tabular}
\end{wraptable}

\begin{figure}[t]
    \centering
    \includegraphics[width=0.65\linewidth]{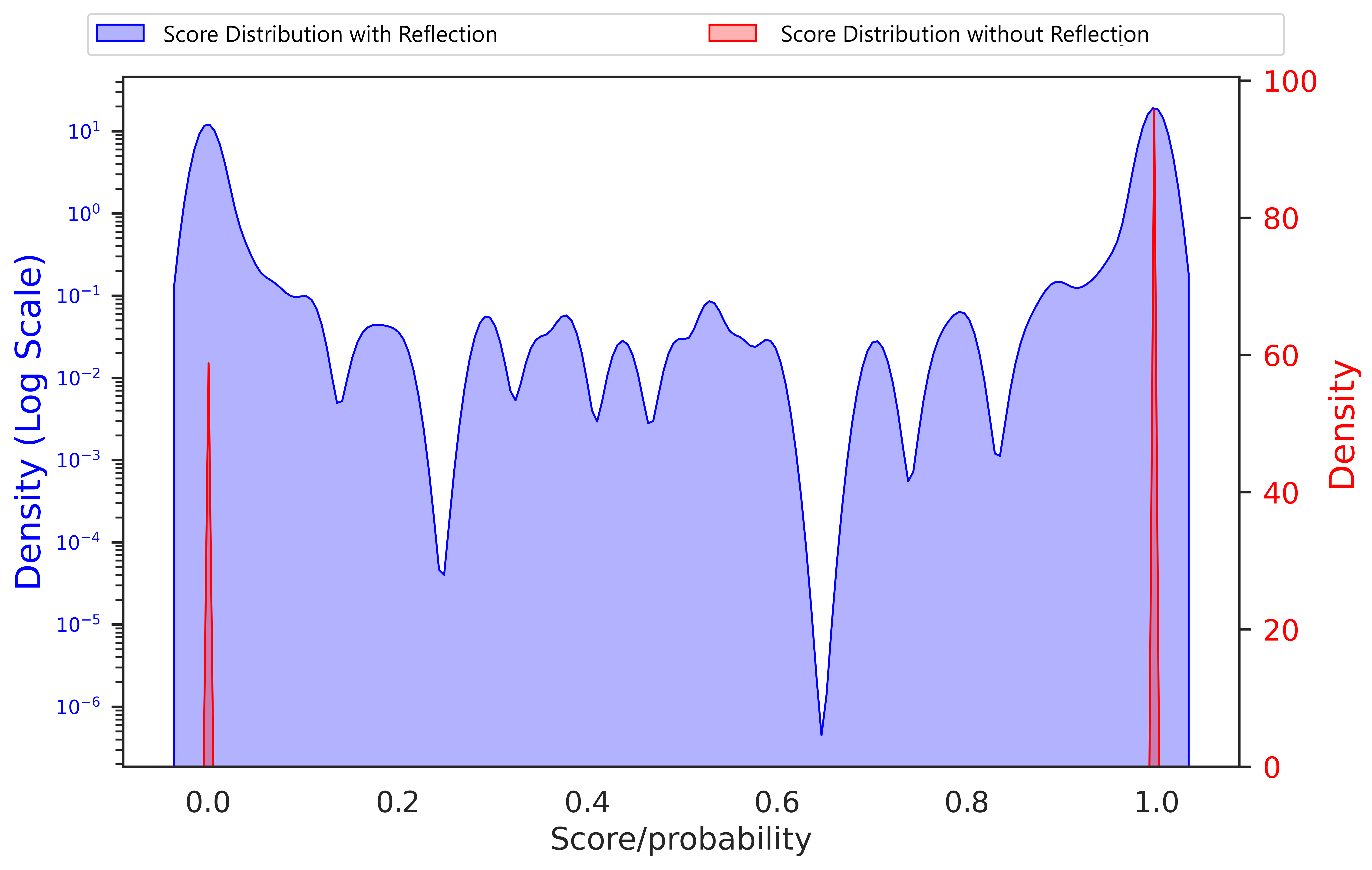}
    \caption{Comparison of score/probability distributions for last- vs.\ first-decision scoring in reflection-aided prompting strategy.}
    \label{fig:first_last_scoring}
\end{figure}

Our experimental results in Table~\ref{tab:first_last_decision_scoring} show that, for the same model, using the \emph{reflection-aided} scoring method yields substantially more stable classification scores than a scoring method that does not incorporate reflection. In addition, as shown in Figure~\ref{fig:first_last_scoring}, the reflection-aided approach produces more calibrated probability distributions, whereas the non-reflective scoring method exhibits highly bimodal behavior that can destabilize threshold-based decision procedures.

\section{Empirical Study of Scaling RL}

\subsection{Data Efficiency}
\begin{figure*}[t]
    \centering
    \begin{subfigure}[t]{0.48\textwidth}
        \centering
        \includegraphics[width=\textwidth]{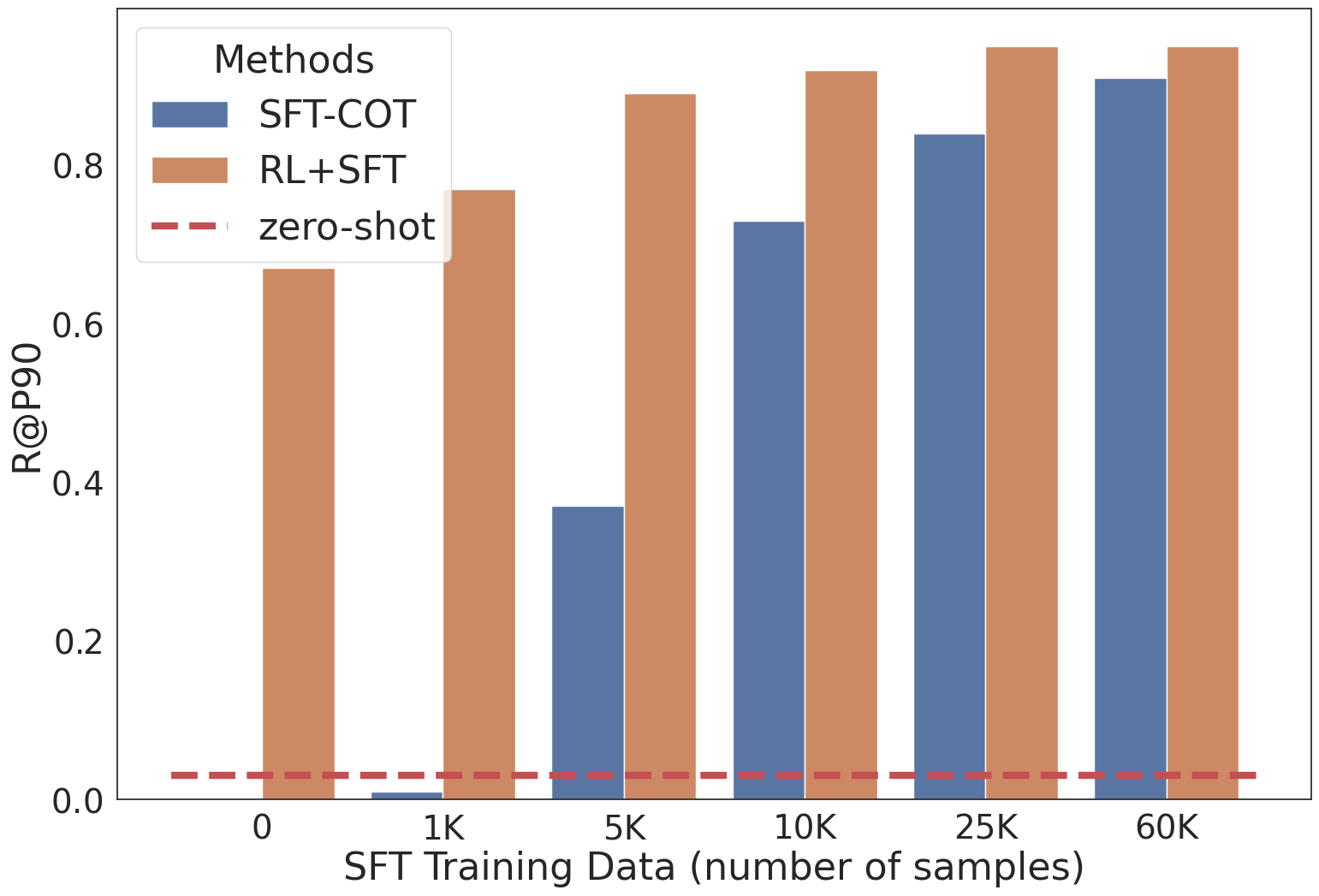}
        \caption{\phishing RL trained using 661 samples}
        \label{fig:sft_data_scaling_a}
    \end{subfigure}
    \hfill
    \begin{subfigure}[t]{0.48\textwidth}
        \centering
        \includegraphics[width=\textwidth]{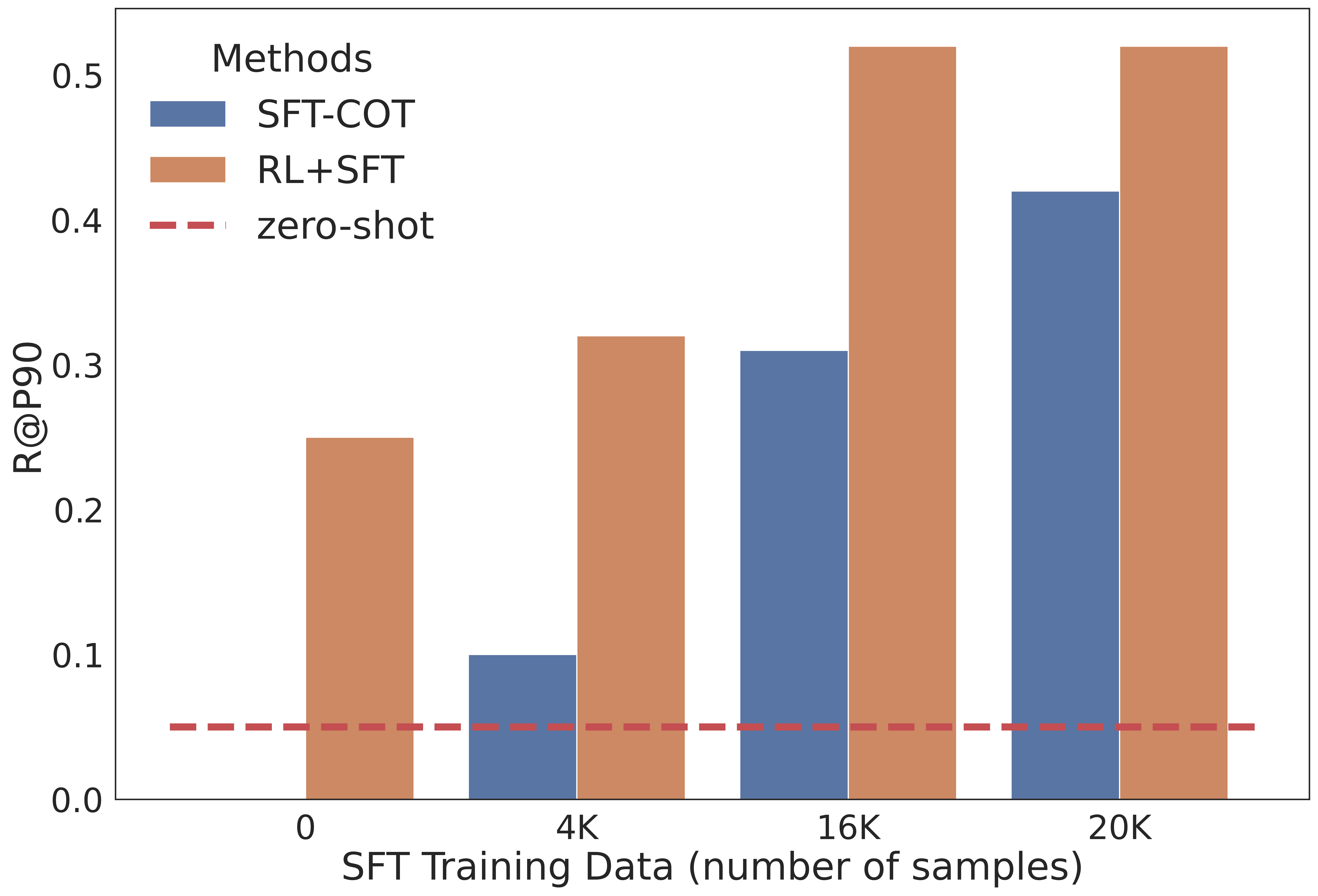}
        \caption{\ubp RL trained using 836 samples}
        \label{fig:sft_data_scaling_b}
    \end{subfigure}
    \par\bigskip
    \begin{subfigure}[t]{0.48\textwidth}
        \centering
        \includegraphics[width=\textwidth]{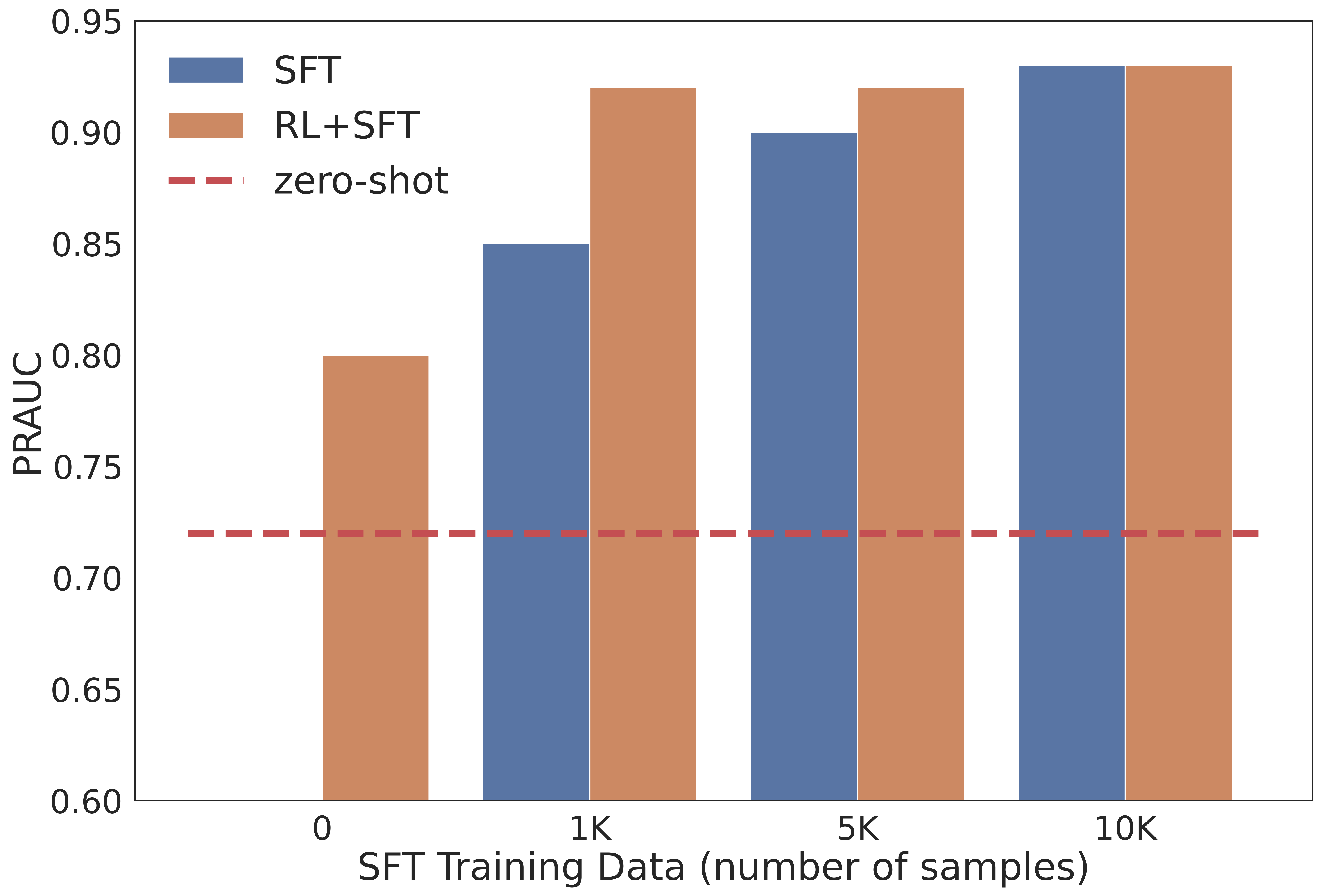}
        \caption{\hpi RL trained using 200 samples}
        \label{fig:sft_data_scaling_c}
    \end{subfigure}    
    \caption{Supervised Fine-tuning data scaling and its impact on Reinforcement learning performance. First bar from the right (SFT data = 0) shows RL-Only training.}
    \label{fig:sft_data_scaling}
\end{figure*}
As shown in Figure~\ref{fig:sft_data_scaling}, we examine how RL improves performance beyond SFT across three representative content moderation tasks—\phishing, \ubp, and \hpi—using the Qwen2.5-VL-7B model. For the RL stage, we employ a small set of high-quality examples sampled in the same manner as our ground-truth standard dataset.

Across tasks, several consistent patterns observed:

\begin{itemize}
    \item \textbf{RL-Only achieves strong performance even with extremely limited data.}  
    When trained on only a few hundred examples, the RL-Only model often matches or surpasses the performance of an SFT model trained on \emph{tens of thousands} of samples. This indicates that RL can be more than an order of magnitude ($\!10\times$) more data-efficient than SFT in these settings.

    \item \textbf{SFT$\rightarrow$RL provides substantial gains when SFT training is moderate.}  
    When the SFT model is trained on thousands of samples, adding an RL stage consistently improves performance across all tasks, typically yielding R@P90 gains of 5--15 percentage points. In this regime, RL effectively corrects residual errors and sharpens decision boundaries.

    \item \textbf{RL gains diminish and eventually saturate as SFT scale increases.}  
    As SFT data grows into the tens of thousands, the performance gap between SFT and SFT$\rightarrow$RL narrows. While large-scale SFT produces a strong initialization, it also constrains exploration by anchoring the policy to learned patterns. Consequently, RL has limited flexibility to discover alternative reasoning paths or higher-quality responses, resulting in diminishing and eventually saturating performance gains.
\end{itemize}

\subsection{Number of training tokens}

\begin{wrapfigure}[15]{r}{0.48\linewidth}
    \vspace{-10pt}
    \centering
    \includegraphics[width=1\linewidth]{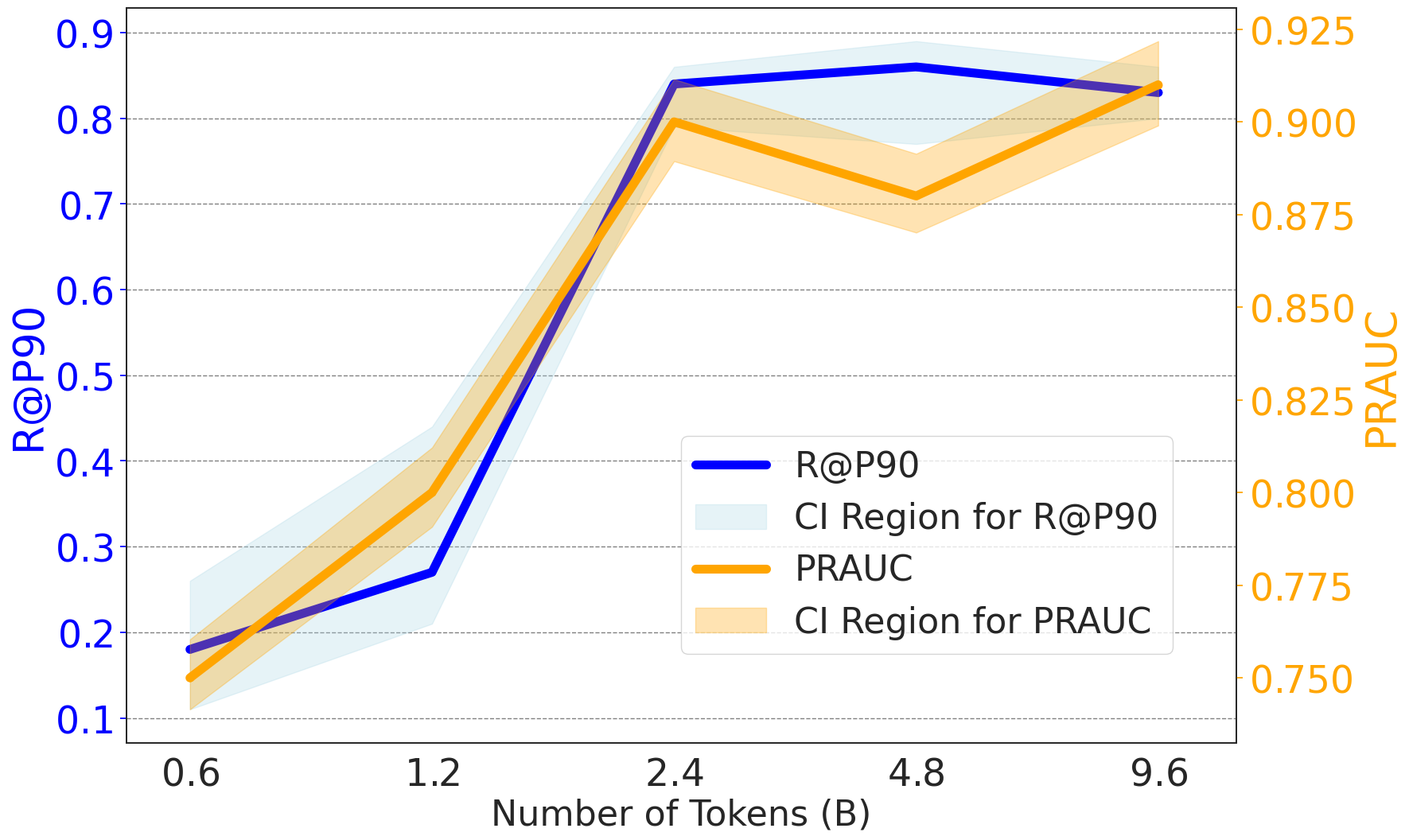}
    \caption{Increasing the data scale in RL training initially boosts model performance, but the gains plateau after a certain threshold.}
    \label{fig:data_scaling}
    \vspace{-50pt}
\end{wrapfigure}

As shown in Figure~\ref{fig:data_scaling}, increasing the number of input tokens per rollout leads to consistent performance gains, but the improvements taper off once the token budget reaches a moderate scale. This behavior follows the characteristic sigmoid-like scaling pattern previously observed in prior work~\citep{khatri2025art}.

\textbf{Low token budgets (0.6B--1.2B):}  
Performance is substantially limited, with both R@P90 and PRAUC remaining low. The wide confidence intervals indicate that the model receives too few comparisons to produce stable gradients.

\textbf{Intermediate token budgets (2.4B):}  
We observe a sharp increase in R@P90 and PRAUC, corresponding to the point at which the model has sufficient context to form reliable comparisons. This regime provides the largest marginal gains.

\textbf{High token budgets (4.8B--9.6B):}  
Performance saturates, and the confidence intervals for R@P90 overlap across settings. Additional tokens provide diminishing returns.

\subsection{Number of Rollouts}
\begin{wraptable}{r}{0.42\linewidth}
\centering
\vspace{-10pt}
\caption{Performance for \phishing and \ubp versus number of rollouts.}
\label{tab:rollouts_phishing_ubp}
\vspace{-10pt}
\begin{tabular}{ccc}
\hline
Number of rollouts & \phishing & \ubp \\
\hline
8   & 0.17& 0.53 \\
16  & 0.18       & 0.56  \\
32  & 0.31       & 0.63  \\
64  & \textbf{0.65} & 0.62 \\
128 & 0.55      & \textbf{0.64 } \\
\hline
\end{tabular}
\vspace{-10pt}
\end{wraptable}
Model performance under GRPO fine-tuning improves as the number of rollouts increases, but the gains diminish and eventually saturate, following an sigmoid-like scaling pattern. Using larger rollout groups leads to more reliable relative comparisons among sampled responses, which in turn produces a cleaner and more stable advantage signal for learning. As a result, increasing the rollout budget—effectively expanding exploration during RL fine-tuning—can improve performance, consistent with prior findings~\citep{hou2025advancing, li2025knapsack}. 

In practice, however, the use of large rollout counts is limited by the capacity of the LLM-based judge used for rubric-based rewards ($R_{\text{rub}}$, Section~\ref{s:reward_shaping}), particularly when multiple rubric criteria must be evaluated in parallel.

\subsubsection{RL improves response selection on rollouts: bound analysis}
Drawing from recent findings \citep{yue2504does, shao2024deepseekmath}, RL often improves performance not by enhancing a model’s underlying reasoning ability, but by increasing the probability of selecting a correct answer from a set of candidate generations. In this view, RL fine-tuning functions primarily as a \emph{response selection} mechanism rather than a fundamental \emph{capability enhancement} step. 

\begin{figure}
    \centering
    \includegraphics[width=0.8\linewidth]{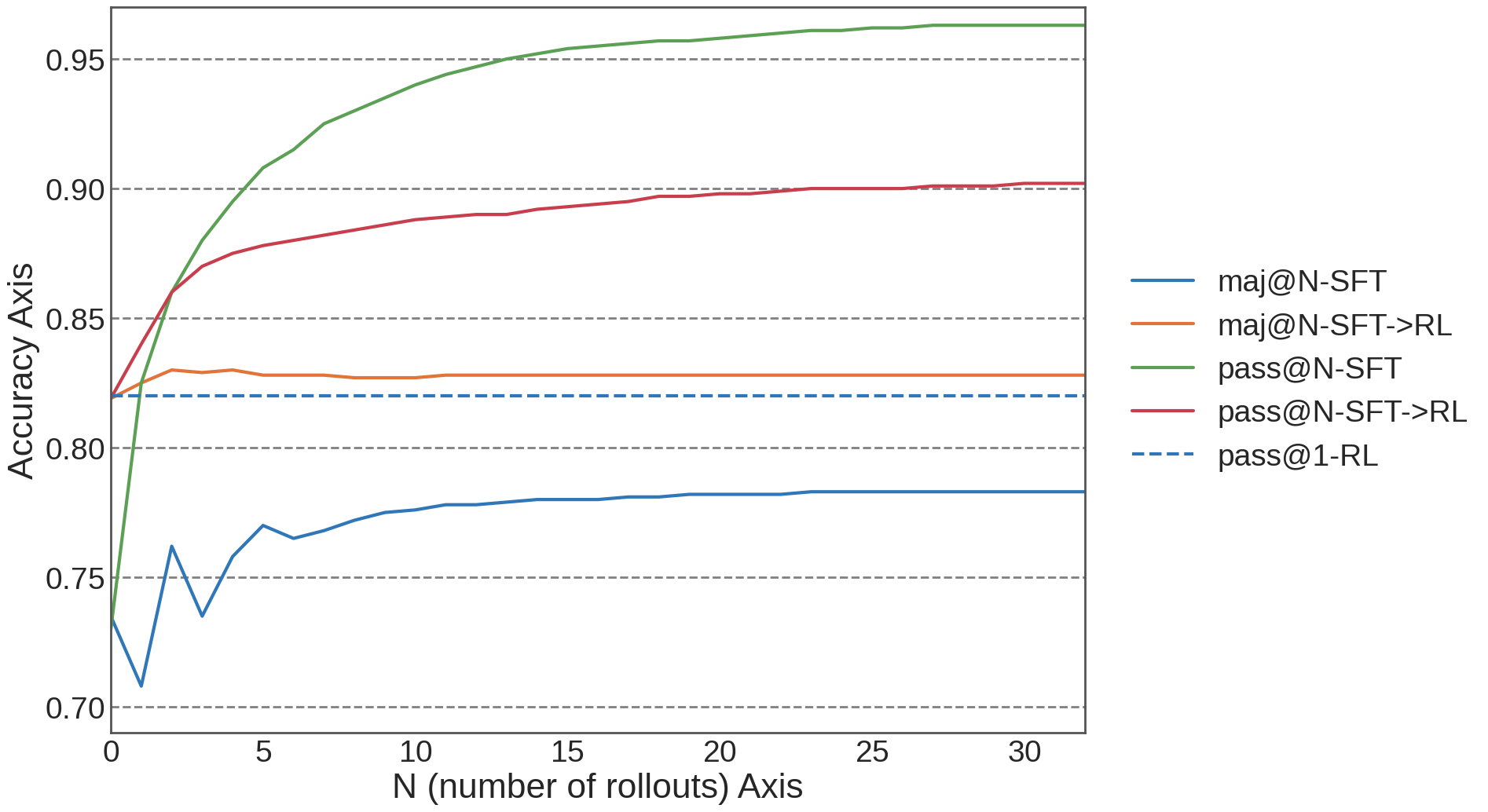}
    \caption{Comparison of accuracy as a function of rollouts \(N\) for SFT and SFT$\rightarrow$RL trained models}
    \label{fig:pass_n_study}

\end{figure}

To examine whether this dynamic also holds for content moderation, we evaluate an SFT baseline and a two-stage SFT$\rightarrow$RL model on the \ubp~dataset. We measure performance using two complementary sampling-based metrics over varying numbers of rollouts ($N \in [1, 32]$):

\textbf{pass@$N$}: The probability that at least one of $N$ independent rollouts yields a correct response.  
    This metric reflects the model’s \emph{best-case correctness} and directly relates to improvements driven by $R_{\text{acc}}$: higher accuracy rewards increase the chance that at least one sampled output is correct.

\textbf{maj@$N$}: The probability that at least half of the $N$ rollouts are correct.  
    This metric captures \emph{consistency} and it complements $R_{\text{acc}}$ by measuring whether RL fine-tuning increases the model’s reliability across samples, not just the probability of producing a single correct output.

\noindent\textbf{Key Observations:}

\begin{enumerate}
\item \textbf{Impact of RL Fine-Tuning:} Following RL training on top of SFT, maj@$N$ demonstrates substantial improvement, increasing from 0.72 to 0.82 at $N=1$, and from 0.77 to 0.83 at $N=32$. This indicates that RL effectively increases the model's propensity to generate correct responses.

\item \textbf{Convergence of Metrics:} The gap between \texttt{pass@N-SFT} and \texttt{maj@N-SFT$\rightarrow$RL} narrows considerably compared to the SFT baseline (\texttt{pass@N-SFT} vs. \texttt{maj@N-SFT}). This convergence suggests that RL training improves output consistency across multiple rollouts, making the model's behavior more deterministic and reliable.

\item \textbf{Performance Ceiling Estimation:} The difference between \texttt{pass@N-SFT} and \texttt{pass@1-SFT} serves as a (loose) upper bound on the potential performance gains achievable through RL-based optimization in two stage SFT$\rightarrow$RL training. If this gap is small, it indicates limited headroom for improvement via better response selection strategies.
\end{enumerate}

\subsection{Reward Shaping}\label{s:reward_shaping}
Taken together, findings in Section~\ref{sec:VerificationAndRewardDesign} illustrate a central tension in RL for LLMs: structural stability and semantic grounding do not naturally co-evolve. To encourage the model to think longer and ground its reasoning, we introduce four rewards. Each reward captures a different aspect of the model's behavior—correctness, reasoning quality, and output structure—providing richer and lower-variance feedback during optimization.

\begin{enumerate}
  \item \textbf{Final Verifiable Accuracy Reward} ($R_{\text{acc}}$):  
  A deterministic reward that checks whether the model's final binary prediction matches the ground-truth label. This reward is fully verifiable and does not require an external judge.

  \item \textbf{Format Reward} ($R_{\text{fmt}}$):  
  Ensures that the model emits both its reasoning trace and final answer in the expected structured format (e.g., reasoning tags, JSON schema, or answer markers).

  \item \textbf{Targeted Reasoning Length Reward} ($R_{\text{len}}$):  
  Encourages outputs to fall within a desired length range, giving the model sufficient ``room for reasoning’’ and preventing collapse into short, label-only responses.

    \item \textbf{Rubric-Based Reasoning Reward} ($R_{\text{rub}}$):  
    To provide supervision beyond the final binary moderation label, we employ a rubric-based reward that evaluates the reasoning trace and reasoning steps. This reward assesses instruction adherence, policy consistency, and the correct application of task-specific criteria. The reasoning decomposes into verifiable checks and the rubric is implemented using either (i) an LLM-as-a-judge or (ii) human-provided rubric annotations. For example, in a profile-matching task, the judge is asked: \emph{“Does the reasoning correctly compare the two profile images according to the rubric?”}. The resulting score is a single scalar reward assigned to the full generation.We employ inverse-frequency weighting to aggregate rewards, ensuring that high-frequency labels do not disproportionately bias the optimization objective.

\end{enumerate}

To improve training stability and avoid reward hacking that often arises with sparse, single-objective rewards, we use a shaped reward that integrates above four rewards signals a single scalar objective:
\begin{equation}
R_{\text{total}} \;=\; 
\alpha_{\text{acc}} \, R_{\text{acc}}
\;+\;
\alpha_{\text{fmt}} \, R_{\text{fmt}}
\;+\;
\alpha_{\text{len}} \, R_{\text{len}}
\;+\;
\alpha_{\text{rub}} \, R_{\text{rub}},
\label{eq:rewards}
\end{equation}
where 
$\alpha_{\text{acc}}$, 
$\alpha_{\text{fmt}}$,
$\alpha_{\text{len}}$,
and
$\alpha_{\text{rub}}$
are non-negative weighting coefficients that satisfy 
$\alpha_{\text{acc}} + \alpha_{\text{fmt}} + \alpha_{\text{len}} + \alpha_{\text{rub}}= 1$, and in our experimentation they are all equally weighted. 
This shaped reward encourages a balanced policy that remains accurate, well-reasoned, and format-consistent throughout training. 

Given the full shaped reward in Eq.~\eqref{eq:rewards}, we evaluate the contribution of each additional component relative to a baseline that uses only the accuracy and format rewards. In our experiments, we observe that the outcome reward score can be quickly reaching high score and prelature, however the rubric score continue to increase for most of the training epochs, indicating model is learning further task structure. We also observe the model can generate more granular reasoning  amid the performance improvement. Two interventions—the targeted reasoning length reward and the rubric-based reasoning reward—produce particularly notable improvements:

\begin{enumerate}
    \item \textbf{Targeted Reasoning Length Reward:}  
    Relative to the accuracy+format baseline, this intervention yields a \(7\%\) improvement in F1 on \textit{Qwen2.5-7B}.

    \item \textbf{Rubric-Based Reasoning Reward:}
    \begin{enumerate}
        \item In \phishing, compared to the accuracy+format+length baseline, applying a rubric-based reward model to the full reasoning trace—focused on faithfulness and factuality—yields a substantial \(12\%\) improvement in F1 (\ref{tab:rub_model_performance}). 
        \item In \hpi compared to the accuracy+format baseline, the rubric-based reasoning reward achieves 4\% higher PRAUC where the rubric labels are verifiable via human annotation. 
    \end{enumerate}
    
\end{enumerate}

\begin{table}[ht]
\centering
\small
\caption{Performance of Qwen models under different reward setups in \phishing. This result highlights the importance of qualitative, rubric-driven supervision for stabilizing learning and improving overall model quality.}
\label{tab:rub_model_performance}
\begin{tabular}{llccc}
\toprule
Model & Reward Setup & Recall & Precision & F1 \\
\midrule
\multirow{3}{*}{Qwen 3 8B}
  & $R_{\text{acc}}+R_{\text{fmt}}$ (Baseline) & 0.58 (0.47, 0.68) & 0.43 (0.34, 0.52) & 0.49 (0.41, 0.58) \\
  & $R_{\text{acc}}+R_{\text{fmt}}+R_{\text{len}}$           & 0.61 (0.48, 0.75) & 0.41 (0.30, 0.54) & 0.49 (0.38, 0.60) \\
  & $R_{\text{acc}}+R_{\text{fmt}}+R_{\text{len}}+R_{\text{rub}}$                & \textbf{0.71} (0.59, 0.83) & \textbf{0.54} (0.43, 0.67) & \textbf{0.61} (0.51, 0.71) \\
\midrule
\multirow{3}{*}{Qwen 2.5 VL 7B}
  & $R_{\text{acc}}+R_{\text{fmt}}$ (Baseline) & 0.49 (0.39, 0.59) & 0.46 (0.37, 0.55) & 0.47 (0.40, 0.55) \\
  & $R_{\text{acc}}+R_{\text{fmt}}+R_{\text{len}}$           & 0.67 (0.55, 0.80) & 0.45 (0.34, 0.57) & 0.54 (0.44, 0.63) \\
  & $R_{\text{acc}}+R_{\text{fmt}}+R_{\text{len}}+R_{\text{rub}}$               & \textbf{0.69} (0.57, 0.81) & \textbf{0.54} (0.42, 0.67) & \textbf{0.60} (0.50, 0.70) \\
\bottomrule
\end{tabular}
\end{table}

\subsection{Effective Batch Size}
\begin{wraptable}{r}{0.48\linewidth}
\vspace{-10pt}
\centering
\caption{\phishing Unweighted R@P90 by effective batch size}
\label{tab:effective-batch-size}
\begin{tabular}{c c}
\hline
Effective \\ batch size & \phishing R@P90 \\
\hline
128  & 0.18 \\
1024 & 0.81 \\
2048 & 0.85 \\
4096 & 0.85 \\
\hline
\end{tabular}
\vspace{-10pt}
\end{wraptable}
The effective batch size plays a critical role in reinforcement learning training stability and convergence. As shown in Table~\ref{tab:effective-batch-size}, increasing the effective batch size from 128 to 1024 dramatically improves performance, with the \phishing detection R@P90 metric increasing from 0.18 to 0.81. Performance plateaus at approximately 0.85 for batch sizes of 2048 and above, suggesting diminishing returns beyond this threshold.
In distributed training frameworks such as TRL and VeRL, the effective batch size is computed as the product of three key components:
\begin{equation}
\text{Effective Batch Size} = B_{\text{local}} \times N_{\text{GPU}} \times N_{\text{accum}}
\label{eq:effective-batch-size}
\end{equation}
where each term represents:
\begin{itemize}
    \item Local batch size ($B_{\text{local}}$): The number of samples processed per GPU device in a single forward pass. This parameter is constrained by GPU memory capacity and depends on model size and input sequence length. In TRL terminology, this is referred to as ``batch size per device,'' while VeRL uses the term ``microbatch size.''
    
    \item Number of GPUs ($N_{\text{GPU}}$): The total number of GPU devices available for distributed training. This represents the available computational resources and enables data parallelism across multiple devices.
    
    \item Gradient accumulation steps ($N_{\text{accum}}$): The number of forward-backward passes performed before updating model parameters. This hyperparameter (denoted as \texttt{gradient\_accumulation\_steps} in most frameworks) requires careful tuning to balance training efficiency and gradient quality.
\end{itemize}
Based on our empirical results, we recommend using an effective batch size of at least 1024 for stable training and optimal performance. This threshold ensures sufficient gradient diversity and reduces variance in policy gradient estimates, which is particularly important for reinforcement learning PPO and GRPO algorithms. 

\section{Disagreement Filtering for Data-Efficient RL}

In this section, we show how the data efficiency of RL can be further improved by leveraging model self-consistency to identify training examples with high learning value. We refer to this approach as \emph{Disagreement Filtering}.

We begin by prompting a pretrained language model (e.g., Qwen3 8B) multiple times for each input to generate diverse reasoning paths and final predictions. We define \emph{disagreement} examples as those for which the sampled predictions do not reach consensus. Among the remaining agreement examples, we further categorize an example as \emph{easy} if all sampled predictions are correct, and \emph{hard} if all predictions are incorrect. Our intuition is that disagreement examples are neither trivially easy nor irreducibly hard, and therefore provide a more informative and stable learning signal for RL optimization.

\begin{table*}[ht]
\centering
\small
\begin{tabular}{lccc}
\toprule
Data Subsets & Data Size & F1 & PRAUC \\
\midrule
All                    & 677 & 0.87 [0.86, 0.89] & 0.85 \\
Disagreement + Easy    & 601 & 0.86 [0.84, 0.87] & 0.84 \\
Easy                   & 566 & 0.79 [0.77, 0.81] & 0.80 \\
Disagreement           &  61 & 0.88 [0.86, 0.89] & 0.90 \\
\bottomrule
\end{tabular}
\caption{Disagreement filtering results on \phishing for Qwen-3 8B. Removing hard or easy examples preserves, and in some cases improves, overall performance despite using substantially fewer training samples.}
\label{tab:negative_sampling}
\end{table*}

We evaluate disagreement-based data filtering on the \phishing task. For each training example, we generate two rollouts at temperatures 0.7, 1.0, and 1.3, resulting in six trajectories per example. Varying the sampling temperature encourages diverse model behaviors and increases the likelihood of uncovering disagreement.

Starting from 677 total examples, this procedure yields 76 hard examples, 61 disagreement examples, and 566 easy examples. We then apply GRPO to different subsets of the dataset, with results summarized in Table~\ref{tab:negative_sampling}.

We observe that removing hard examples consistently improves performance relative to training on the full dataset. While counter-intuitive at first glance, this result suggests that hard examples may introduce noisy or unstable reward signals, leading to overfitting or suboptimal policy updates. In contrast, GRPO trained on a smaller but more carefully curated dataset achieves comparable—and in some cases superior—performance.

Importantly, these gains translate into substantial improvements in data efficiency. As shown in Figure~\ref{fig:sft_data_scaling}, RL already achieves approximately \(10\times\) higher data efficiency than SFT. When combined with disagreement filtering, RL trained on only the disagreement subset (61 examples) attains performance comparable to SFT trained on the full dataset, corresponding to an effective \(100\times\) improvement in data efficiency. 

Overall, these findings demonstrate that selecting training data based on model disagreement and estimated difficulty is a powerful and practical strategy for improving the sample efficiency and stability of RL-based content moderation systems.

\section{Conclusion}
In this work, we present a systematic empirical study of scaling reinforcement learning (RL) for large language model–based content moderation, a domain characterized by label scarcity, evolving policies, and high demands for nuanced, policy-grounded reasoning. Across three real-world moderation tasks, we show that RL enables general-purpose LLMs to be transformed into specialized classifiers that substantially outperform supervised fine-tuning (SFT) under limited data regimes. Our results demonstrate that RL follows predictable, sigmoid-like scaling behavior with respect to data, rollouts, and compute, providing practical guidance for allocating resources in industrial moderation pipelines. Critically, RL achieves up to one to two orders of magnitude higher data efficiency than SFT, making it particularly well suited for domains where expert annotations are expensive or slow to obtain.

We further identify and address key failure modes that arise when applying RL to content moderation, including reward hacking, reasoning-length collapse, bimodal confidence distributions, and trade-offs between faithfulness and factuality. Through a combination of reward shaping, rubric-based reasoning rewards, Monte-Carlo score aggregation, and reflection-aided prompting, we show that these issues can be substantially mitigated in practice. Together, these techniques yield stable training dynamics, better-calibrated confidence estimates, and more reliable policy-grounded reasoning. Our findings suggest that RL, when carefully designed and scaled, offers a principled and effective path toward building robust, expert-level content moderation systems capable of adapting to complex and evolving policy requirements.

\clearpage
\newpage
\bibliographystyle{assets/plainnat}
\bibliography{custom}

@article{wei2022chain,
  title={Chain-of-thought prompting elicits reasoning in large language models},
  author={Wei, Jason and Wang, Xuezhi and Schuurmans, Dale and Bosma, Maarten and Xia, Fei and Chi, Ed and Le, Quoc V and Zhou, Denny and others},
  journal={Advances in neural information processing systems},
  volume={35},
  pages={24824--24837},
  year={2022}
}

@article{mu2024rule,
  title={Rule based rewards for language model safety},
  author={Mu, Tong and Helyar, Alec and Heidecke, Johannes and Achiam, Joshua and Vallone, Andrea and Kivlichan, Ian and Lin, Molly and Beutel, Alex and Schulman, John and Weng, Lilian},
  journal={Advances in Neural Information Processing Systems},
  volume={37},
  pages={108877--108901},
  year={2024}
}

@inproceedings{mu2024ruleicml,
  title={Rule based rewards for fine-grained llm safety},
  author={Mu, Tong and Helyar, Alec and Heidecke, Johannes and Achiam, Joshua and Vallone, Andrea and Kivlichan, Ian D and Lin, Molly and Beutel, Alex and Schulman, John and Weng, Lilian},
  booktitle={ICML 2024 Next Generation of AI Safety Workshop},
  year={2024}
}

@misc{markov2024systems,
  title={Systems and methods for language model-based content classification},
  author={MARKOV, Todor and Zhang, Chong and AGARWAL, Sandhini and NEKOUL, Florentine Mary ELOUNDOU and Lee, Theodore and Adler, Steven and Jiang, Angela and WENG, Lilian},
  year={2024},
  month=oct # "~31",
  publisher={Google Patents},
  note={US Patent App. 18/308,586}
}

@article{yuan2025hard,
  title={From hard refusals to safe-completions: Toward output-centric safety training},
  author={Yuan, Yuan and Sriskandarajah, Tina and Brakman, Anna-Luisa and Helyar, Alec and Beutel, Alex and Vallone, Andrea and Jain, Saachi},
  journal={arXiv preprint arXiv:2508.09224},
  year={2025}
}

@article{yuan2024rigorllm,
  title={Rigorllm: Resilient guardrails for large language models against undesired content},
  author={Yuan, Zhuowen and Xiong, Zidi and Zeng, Yi and Yu, Ning and Jia, Ruoxi and Song, Dawn and Li, Bo},
  journal={arXiv preprint arXiv:2403.13031},
  year={2024}
}

@inproceedings{kumar2024watch,
  title={Watch your language: Investigating content moderation with large language models},
  author={Kumar, Deepak and AbuHashem, Yousef Anees and Durumeric, Zakir},
  booktitle={Proceedings of the International AAAI Conference on Web and Social Media},
  volume={18},
  pages={865--878},
  year={2024}
}

@misc{ma2024adaptinglargelanguagemodels,
      title={Adapting Large Language Models for Content Moderation: Pitfalls in Data Engineering and Supervised Fine-tuning}, 
      author={Huan Ma and Changqing Zhang and Huazhu Fu and Peilin Zhao and Bingzhe Wu},
      year={2024},
      eprint={2310.03400},
      archivePrefix={arXiv},
      primaryClass={cs.LG},
      url={https://arxiv.org/abs/2310.03400}, 
}

@misc{vonwerra2020trl,
  title = {Transformers Reinforcement Learning (TRL)},
  author = {Leandro von Werra and others},
  year = {2020},
  url = {https://github.com/huggingface/trl}
}

@misc{verl2024hybridflow,
  title = {Verl: Efficient Reinforcement Learning for LLMs},
  author = {Verl-Team},
  year = {2024},
  url = {https://github.com/verlml/verl}
}

@article{liu2025understanding,
  title={Understanding r1-zero-like training: A critical perspective},
  author={Liu, Zichen and Chen, Changyu and Li, Wenjun and Qi, Penghui and Pang, Tianyu and Du, Chao and Lee, Wee Sun and Lin, Min},
  journal={arXiv preprint arXiv:2503.20783},
  year={2025}
}

@inproceedings{muennighoff2025s1,
  title={s1: Simple test-time scaling},
  author={Muennighoff, Niklas and Yang, Zitong and Shi, Weijia and Li, Xiang Lisa and Fei-Fei, Li and Hajishirzi, Hannaneh and Zettlemoyer, Luke and Liang, Percy and Cand{\`e}s, Emmanuel and Hashimoto, Tatsunori B},
  booktitle={Proceedings of the 2025 Conference on Empirical Methods in Natural Language Processing},
  pages={20286--20332},
  year={2025}
}

@article{zheng2025group,
  title={Group sequence policy optimization},
  author={Zheng, Chujie and Liu, Shixuan and Li, Mingze and Chen, Xiong-Hui and Yu, Bowen and Gao, Chang and Dang, Kai and Liu, Yuqiong and Men, Rui and Yang, An and others},
  journal={arXiv preprint arXiv:2507.18071},
  year={2025}
}

@article{shao2024deepseekmath,
  title={Deepseekmath: Pushing the limits of mathematical reasoning in open language models},
  author={Shao, Zhihong and Wang, Peiyi and Zhu, Qihao and Xu, Runxin and Song, Junxiao and Bi, Xiao and Zhang, Haowei and Zhang, Mingchuan and Li, YK and Wu, Yang and others},
  journal={arXiv preprint arXiv:2402.03300},
  year={2024}
}

@article{schulman2017proximal,
  title={Proximal policy optimization algorithms},
  author={Schulman, John and Wolski, Filip and Dhariwal, Prafulla and Radford, Alec and Klimov, Oleg},
  journal={arXiv preprint arXiv:1707.06347},
  year={2017}
}

@inproceedings{qiao2024scaling,
  title={Scaling Up LLM Reviews for Google Ads Content Moderation},
  author={Qiao, Wei and Dogra, Tushar and Stretcu, Otilia and Lyu, Yu-Han and Fang, Tiantian and Kwon, Dongjin and Lu, Chun-Ta and Luo, Enming and Wang, Yuan and Chia, Chih-Chun and others},
  booktitle={WSDM},
  year={2024}
}

@article{kiela2020hateful,
  title={The hateful memes challenge: Detecting hate speech in multimodal memes},
  author={Kiela, Douwe and Firooz, Hamed and Mohan, Aravind and Goswami, Vedanuj and Singh, Amanpreet and Ringshia, Pratik and Testuggine, Davide},
  journal={Advances in neural information processing systems},
  volume={33},
  pages={2611--2624},
  year={2020}
}

@article{yin2025bingoguard,
  title={Bingoguard: Llm content moderation tools with risk levels},
  author={Yin, Fan and Laban, Philippe and Peng, Xiangyu and Zhou, Yilun and Mao, Yixin and Vats, Vaibhav and Ross, Linnea and Agarwal, Divyansh and Xiong, Caiming and Wu, Chien-Sheng},
  journal={arXiv preprint arXiv:2503.06550},
  year={2025}
}

@inproceedings{alam2022survey,
  title={A survey on multimodal disinformation detection},
  author={Alam, Firoj and Cresci, Stefano and Chakraborty, Tanmoy and Silvestri, Fabrizio and Dimitrov, Dimiter and Da San Martino, Giovanni and Shaar, Shaden and Firooz, Hamed and Nakov, Preslav},
  booktitle={Proceedings of the 29th international conference on computational linguistics},
  pages={6625--6643},
  year={2022}
}

@article{yue2504does,
  title={Does reinforcement learning really incentivize reasoning capacity in llms beyond the base model?},
  author={Yue, Yang and Chen, Zhiqi and Lu, Rui and Zhao, Andrew and Wang, Zhaokai and Song, Shiji and Huang, Gao},
  journal={arXiv preprint arXiv:2504.13837},
  year={2025}
}

@article{inan2023llamaguard,
  title={Llama Guard: LLM-based Input-Output Safeguard for Human-AI Conversations},
  author={Inan, Huseyin and Upasani, Kartik and Chi, John and others},
  journal={arXiv preprint arXiv:2312.06674},
  year={2023}
}

@article{sharma2025constitutional,
  title={Constitutional classifiers: Defending against universal jailbreaks across thousands of hours of red teaming},
  author={Sharma, Mrinank and Tong, Meg and Mu, Jesse and Wei, Jerry and Kruthoff, Jorrit and Goodfriend, Scott and Ong, Euan and Peng, Alwin and Agarwal, Raj and Anil, Cem and others},
  journal={arXiv preprint arXiv:2501.18837},
  year={2025}
}

@misc{openai_moderation,
  title={OpenAI Moderation System Guide: Identify potentially harmful content in text and images.},
  author={OpenAI-Platform},
  year={2023},
  note={https://platform.openai.com/docs/guides/moderation}
}

@inproceedings{anthropic_constitutional_ai,
  title={Constitutional AI: Harmlessness from AI Feedback},
  author={Bai, Yuntao and Kadavath, S. and others},
  booktitle={NeurIPS},
  year={2022}
}

@article{dai2023safe,
  title={Safe rlhf: Safe reinforcement learning from human feedback},
  author={Dai, Josef and Pan, Xuehai and Sun, Ruiyang and Ji, Jiaming and Xu, Xinbo and Liu, Mickel and Wang, Yizhou and Yang, Yaodong},
  journal={arXiv preprint arXiv:2310.12773},
  year={2023}
}

@article{guan2024deliberative,
  title={Deliberative alignment: Reasoning enables safer language models},
  author={Guan, Melody Y and Joglekar, Manas and Wallace, Eric and Jain, Saachi and Barak, Boaz and Helyar, Alec and Dias, Rachel and Vallone, Andrea and Ren, Hongyu and Wei, Jason and others},
  journal={arXiv preprint arXiv:2412.16339},
  year={2024}
}

@article{zhang2025realsafe,
  title={Realsafe-r1: Safety-aligned deepseek-r1 without compromising reasoning capability},
  author={Zhang, Yichi and Zeng, Zihao and Li, Dongbai and Huang, Yao and Deng, Zhijie and Dong, Yinpeng},
  journal={arXiv preprint arXiv:2504.10081},
  year={2025}
}

@misc{meta_connect2024_responsible_approach,
  title        = {Connect 2024: The responsible approach we are taking to generative AI},
  author       = {{Meta Platforms}},
  year         = {2024},
  howpublished = {\url{https://ai.meta.com/blog/responsible-ai-connect-2024/}},
}

@misc{google_text_moderation_2023,
  title        = {Google Cloud Text Moderation},
  author       = {{Google Cloud}},
  year         = {2023},
  howpublished = {\url{https://cloud.google.com/blog/products/ai-machine-learning/google-cloud-text-moderation}},
}

@inproceedings{markov2023holistic,
  title={A holistic approach to undesired content detection in the real world},
  author={Markov, Todor and Zhang, Chong and Agarwal, Sandhini and Nekoul, Florentine Eloundou and Lee, Theodore and Adler, Steven and Jiang, Angela and Weng, Lilian},
  booktitle={Proceedings of the AAAI conference on artificial intelligence},
  volume={37},
  pages={15009--15018},
  year={2023}
}

@article{weidinger2024holistic,
  title={Holistic safety and responsibility evaluations of advanced ai models},
  author={Weidinger, Laura and Barnhart, Joslyn and Brennan, Jenny and Butterfield, Christina and Young, Susie and Hawkins, Will and Hendricks, Lisa Anne and Comanescu, Ramona and Chang, Oscar and Rodriguez, Mikel and others},
  journal={arXiv preprint arXiv:2404.14068},
  year={2024}
}

@inproceedings{gampa2023prioritised,
  title={Prioritised Moderation for Online Advertising},
  author={Gampa, Phanideep and Valsangkar, Akash Anil and Choubey, Shailesh and others},
  booktitle={Proceedings of the IEEE/CVF Conference on Computer Vision and Pattern Recognition},
  pages={2004--2012},
  year={2023}
}

@article{khatri2025art,
  title={The art of scaling reinforcement learning compute for llms},
  author={Khatri, Devvrit and Madaan, Lovish and Tiwari, Rishabh and Bansal, Rachit and Duvvuri, Sai Surya and Zaheer, Manzil and Dhillon, Inderjit S and Brandfonbrener, David and Agarwal, Rishabh},
  journal={arXiv preprint arXiv:2510.13786},
  year={2025}
}

@article{huang2024survey,
  title={A Survey on Hallucination in Large Language Models: Principles, Taxonomy, Challenges, and Open Questions},
  author={Huang, Lei and Yu, Weijiang and Ma, Weitao and Zhong, Weihong and Feng, Zhangyin and Wang, Haotian and Chen, Qianglong and Peng, Weihua and Feng, Xiaocheng and Qin, Bing and Liu, Ting},
  journal={ACM Transactions on Information Systems},
  volume={42},
  number={1},
  year={2024},
  publisher={Association for Computing Machinery},
  url={https://doi.org/10.1145/3703155}
}

@article{hughes2023hhem,
  title = {Cut the Bull... Detecting Hallucinations in Large Language Models},
  author = {Hughes, Simon},
  journal = {Vectara Blog},
  year = {2023},
  month = {November},
  url = {https://www.vectara.com/blog/cut-the-bull-detecting-hallucinations-in-large-language-models}
}

@article{hou2025advancing,
  title={Advancing language model reasoning through reinforcement learning and inference scaling},
  author={Hou, Zhenyu and Lv, Xin and Lu, Rui and Zhang, Jiajie and Li, Yujiang and Yao, Zijun and Li, Juanzi and Tang, Jie and Dong, Yuxiao},
  journal={arXiv preprint arXiv:2501.11651},
  year={2025}
}

@article{li2025knapsack,
  title={Knapsack rl: Unlocking exploration of llms via optimizing budget allocation},
  author={Li, Ziniu and Chen, Congliang and Yang, Tianyun and Ding, Tian and Sun, Ruoyu and Zhang, Ge and Huang, Wenhao and Luo, Zhi-Quan},
  journal={arXiv preprint arXiv:2509.25849},
  year={2025}
}

@article{shinn2023reflexion,
  title={Reflexion: Language agents with verbal reinforcement learning, 2023},
  author={Shinn, Noah and Cassano, Federico and Labash, Beck and Gopinath, Ashwin and Narasimhan, Karthik and Yao, Shunyu},
  journal={URL https://arxiv. org/abs/2303.11366},
  volume={1},
  year={2023}
}

@article{gao2025cannot,
  title={" I Cannot Write This Because It Violates Our Content Policy": Understanding Content Moderation Policies and User Experiences in Generative AI Products},
  author={Gao, Lan and Chen, Oscar and Lee, Rachel and Feamster, Nick and Tan, Chenhao and Chetty, Marshini},
  journal={arXiv preprint arXiv:2506.14018},
  year={2025}
}

@article{comanici2025gemini,
  title={Gemini 2.5: Pushing the frontier with advanced reasoning, multimodality, long context, and next generation agentic capabilities},
  author={Comanici, Gheorghe and Bieber, Eric and Schaekermann, Mike and Pasupat, Ice and Sachdeva, Noveen and Dhillon, Inderjit and Blistein, Marcel and Ram, Ori and Zhang, Dan and Rosen, Evan and others},
  journal={arXiv preprint arXiv:2507.06261},
  year={2025}
}

@article{sharma2022detecting,
  title={Detecting and understanding harmful memes: A survey},
  author={Sharma, Shivam and Alam, Firoj and Akhtar, Md Shad and Dimitrov, Dimitar and Martino, Giovanni Da San and Firooz, Hamed and Halevy, Alon and Silvestri, Fabrizio and Nakov, Preslav and Chakraborty, Tanmoy},
  journal={arXiv preprint arXiv:2205.04274},
  year={2022}
}

@article{cobbe2021training,
  title={Training verifiers to solve math word problems},
  author={Cobbe, Karl and Kosaraju, Vineet and Bavarian, Mohammad and Chen, Mark and Jun, Heewoo and Kaiser, Lukasz and Plappert, Matthias and Tworek, Jerry and Hilton, Jacob and Nakano, Reiichiro and others},
  journal={arXiv preprint arXiv:2110.14168},
  year={2021}
}

@article{jha2024rlsf,
  title={RLSF: Fine-tuning LLMs via Symbolic Feedback},
  author={Jha, Piyush and Jana, Prithwish and Suresh, Pranavkrishna and Arora, Arnav and Ganesh, Vijay},
  journal={arXiv preprint arXiv:2405.16661},
  year={2024}
}



\end{document}